\documentclass[sigconf]{acmart}

\usepackage{amsfonts}       
\usepackage{amsmath}
\usepackage{amssymb}
\usepackage{graphicx, subfig}
\usepackage{nicefrac} 
\usepackage[ruled, vlined, linesnumbered]{algorithm2e}
\usepackage{enumerate}
\usepackage{enumitem}
\usepackage{geometry}
\usepackage{tikz}
\usetikzlibrary{shapes.geometric}
\usepackage{arydshln}
\usepackage{xspace}
\usepackage{amssymb}
\usepackage{bbding}
\usepackage{tcolorbox}
\usepackage{multirow}
\usepackage{threeparttable}
\usepackage{soul}
\usepackage{color}

\usepackage[capitalize,noabbrev]{cleveref}
\usepackage{makecell}
\usepackage{wrapfig}
\usepackage{caption}
\usepackage{subcaption}

\AtBeginDocument{%
  \providecommand\BibTeX{{%
    \normalfont B\kern-0.5em{\scshape i\kern-0.25em b}\kern-0.8em\TeX}}}

\copyrightyear{2024}
\acmYear{2024}
\setcopyright{acmlicensed}
\acmConference[CIKM '24] {Proceedings of the 33rd ACM International Conference on Information and Knowledge Management}{October 21--25, 2024}{Boise, ID, USA.}
\acmBooktitle{Proceedings of the 33rd ACM International Conference on Information and Knowledge Management (CIKM '24), October 21--25, 2024, Boise, ID, USA}
\acmISBN{979-8-4007-0436-9/24/10}
\acmDOI{10.1145/3627673.3679776}

\begin{document}
\title{Beyond Over-smoothing: Uncovering the Trainability Challenges in Deep Graph Neural Networks}

\author{Jie Peng}
\affiliation{
  \institution{Renmin University of China}
  \city{Beijing}\country{China}
  }
\email{peng_jie@ruc.edu.cn}

\author{Runlin Lei}
\affiliation{
  \institution{Renmin University of China}
  \city{Beijing}\country{China}
  }
\email{runlin_lei@ruc.edu.cn}

\author{Zhewei Wei}
\authornote{Zhewei Wei is the corresponding author. The work was partially done at Gaoling School of Artificial Intelligence, Beijing Key Laboratory of Big Data Management and Analysis Methods, MOE Key Lab of Data Engineering and Knowledge Engineering, and Pazhou Laboratory (Huangpu), Guangzhou, Guangdong 510555, China.
}
\affiliation{
  \institution{Renmin University of China}
  \city{Beijing}\country{China}
  }
\email{zhewei@ruc.edu.cn}

\begin{abstract}

The drastic performance degradation of Graph Neural Networks (GNNs) as the depth of the graph propagation layers exceeds 8-10 is widely attributed to a phenomenon of \emph{Over-smoothing}.
Although recent research suggests that Over-smoothing may not be the dominant reason for such a performance degradation, they have not provided rigorous analysis from a theoretical view, which warrants further investigation.
In this paper, we systematically analyze the real dominant problem in deep GNNs and identify the issues that these GNNs towards addressing Over-smoothing essentially work on via empirical experiments and theoretical gradient analysis. 
We theoretically prove that \emph{the difficult training problem of deep MLPs} is actually the main challenge, and various existing methods that supposedly tackle Over-smoothing actually improve the trainability of MLPs, which is the main reason for their performance gains. 
Our further investigation into trainability issues reveals that properly constrained smaller upper bounds of gradient flow notably enhance the trainability of GNNs. 
Experimental results on diverse datasets demonstrate consistency between our theoretical findings and empirical evidence.
Our analysis provides new insights in constructing deep graph models.

\end{abstract}



\maketitle

\section{Introduction}
Graph Neural Networks (GNNs) have shown remarkable performance in various fields, such as social networks~\cite{defferrard2016convolutional,fan2019graph,zheng2024survey}, molecular discovery~\cite{gilmer2017neural,satorras2021n,wang2022molecular}, computer vision~\cite{zhao2019semantic,han2023vision,pradhyumna2021graph}, $\emph{etc}$. 
Despite the great success of GNNs, the performances of several GNNs degrade notably with increased graph propagation layer depth~\cite{kipf2016semi}. 
It is universally acknowledged that Over-smoothing is the main factor that leads to such a performance drop~\cite{li2018deeper,giraldo2023trade}.

Over-smoothing occurs when the output node embeddings tend to be highly identical after stacking multiple layers of GNNs~\cite{li2018deeper,zhao2019pairnorm,oono2019graph}.
However, even shallow GCN~\cite{kipf2016semi} exhibits a notable performance degradation at around 8 layers, while the smoothing rate of graph propagation is not particularly rapid, which could be demonstrated from the perspectives of Dirichlet Energy~\cite{zhou2021dirichlet}, lazy random walk~\cite{chen2020simple}, Markov chains~\cite{levin2017markov}.
Thus, the presumed strong correlation between Over-smoothing and performance degradation may be {overstated}. 
Although many attempts have been made to analyze that Over-smoothing is not the main cause of performance drop in deep GNNs~\cite{zhou2020effective,zhang2022model,luan2023training}, they have not provided rigorous analysis from a theoretical view, suggesting that the underlying dominant problem of deep GNNs warrants \textbf{further theoretical study}.

Recently, based on the dubious conclusion that Over-smoothing is the dominant problem in deep GNNs, plenty of methods have been put forward to tackle Over-smoothing mainly from the perspectives of Residual Connection~\cite{li2019deepgcns,chen2020simple}, Normalization~\cite{ioffe2015batch,zhao2019pairnorm,zhou2020towards}, and Drop Operation~\cite{rong2019dropedge,feng2020graph}, $\emph{etc}$. 
In these newly proposed deep models targeting Over-smoothing, their performances vary widely. 
For instance, deep GCNII~\cite{chen2020simple} exhibits remarkably good model performance, while deep GCN(DropEdge)~\cite{rong2019dropedge} and GCN(BatchNorm)~\cite{ioffe2015batch,zhou2020towards} still witness poor performance (see~\cref{tab:acc-bn-de-ii-new}) under deep model architecture.
The disparity in the performance among these models may not be solely attributed to their specific designs. 
Instead, it is more likely to depend on whether they focus on effectively addressing the {actual dominant problem} in deep GNNs, where \textbf{attribution errors} may exist in previous solutions.

Naturally, two clear questions arise: 
$\textbf{Q1.}$ \emph{What is the real dominant problem of deep GNNs that leads to the performance drop?} 
$\textbf{Q2.}$ \emph{What problem do these models towards solving Over-smoothing really work on? And what is the key to their performance gains?}

In this paper, we systematically analyze $\textbf{Q1}$ and $\textbf{Q2}$ from both empirical and theoretical perspectives. 

For $\textbf{Q1}$, we analyze the convergence rate of smoothing and evaluate the respective influence of the graph propagation process and the training process through a series of \textbf{decoupled experiments} and novel \textbf{gradient derivations}. 
As a result, our empirical and theoretical analysis indicates that Over-smoothing is not the dominant problem of deep GNNs, and the difficult training problem of deep Multilayer Perceptrons (MLPs) actually has a greater influence, which fills a theoretical gap in previous works.

For $\textbf{Q2}$, we revisit GCNII~\cite{chen2020simple}, Batch Normalization~\cite{ioffe2015batch}, and DropEdge~\cite{rong2019dropedge}, three representative methods of Residual Connection, Normalization, and Drop Operation.
Our goal is to identify the problems they essentially work on.
The results of our decoupled experiments and gradient analysis novelly show that certain modified GNNs like GCN(DropEdge) still perform poorly with increased layer depth~\cite{rong2019dropedge} since they merely solve \textbf{the side issue}, Over-smoothing. 
In contrast, GNNs like GCNII perform well since they effectively address \textbf{the dominant issue}, the trainability problem. 

After clarifying the importance of trainability, we further theoretically explore the gradient of deep GNNs during training by analyzing the order of the gradient flow upper bounds. 
Our findings illustrate that a more stable gradient flow (lower gradient flow upper bound) brought by a \textbf{properly constrained smaller upper bound} of the gradient is one of the key reasons for the good trainability of deep GNNs. 
For instance, the techniques of GCNII effectively enhance trainability by regulating the gradient upper bound to a relatively small and proper level.
We conduct extensive experiments on various datasets to verify the correctness of our theory.
Finally, we have a discussion on the future development of deep GNNs, that is, effectively solving the real dominant problem of deep GNNs, the trainability issue, via regulating the gradient upper bound or adjusting the current paradigm of graph models .

Our main contributions are summarized as follows:
\begin{itemize}
    \item \textbf{Trainability of deep GNNs is more of a problem than Over-smoothing.} 
    We thoroughly analyze Over-smoothing, an overstated problem of deep GNNs, from empirical and theoretical views. Our novel theoretical analysis proves that the difficult training problem of deep MLPs is the dominant problem, which fills a theoretical gap in previous works.
    \item \textbf{The performance gain is mainly due to the effective solution of trainability issues.} 
    We novelly find that various GNNs towards solving Over-smoothing from the views of Residual Connection, Normalization, and Drop Operation actually work on different problems (Over-smoothing or the trainability issue) and result in varying model performances. 
    Whether the training problem is solved effectively is the key to the performance gain of deep GNNs.
    \item \textbf{More stable gradient flow, higher trainability, better performance.} 
    We further explore the gradient flow upper bound in a theoretical way. The order of the upper bounds for various deep GNNs reflects the stability of the gradient flow, which is highly related to the trainability and performance of the models. 
    Experimental results demonstrate consistency between our theoretical findings and empirical evidence.
\end{itemize}

\section{Problem Statement}

\textbf{Notations.}~~We consider an undirected graph $G=(V,E)$ with $v$ nodes and $e$ edges. 
The degree of the $i$-th node is given as $d_i$ and its 1-neighborhood is defined as $\mathcal{N}_i$.
We denote the adjacency matrix of $G$ as $\mathbf{A}\in\mathbb{R}^{v\times v}$. 
Let $\mathbf{D}\in\mathbb{R}^{v\times v}$ denote the diagonal degree matrix. 
The adjacency matrix of $\widehat{G}$ with self-loops is defined as $\mathbf{\widehat{A}}=\mathbf{A}+\mathbf{I}$, and $\mathbf{\widehat{D}}$ represents the corresponding diagonal degree matrix. 
The normalized Laplacian matrix is given as $\mathbf{L}_{sym}=\mathbf{I}-\widehat{\mathbf{D}}^{-\frac{1}{2}}\widehat{\mathbf{A}}\widehat{\mathbf{D}}^{-\frac{1}{2}}$ with eigendecomposition $\mathbf{U}\mathbf{\Lambda}\mathbf{U^{\top}}$, where $\mathbf{U}\in\mathbb{R}^{v\times v}$ consists of the eigenvectors ($u_1$, $u_2$, ... , $u_v$) of $\mathbf{L}_{sym}$.
The eigenvalues of $\mathbf{L}_{sym}$, the elements of diagonal matrix $\mathbf{\Lambda}$, satisfy: $0=a_1\leq a_2\leq...\leq a_v\leq 2$ ($a_v=2$ if and only if $G$ is bipartite).
We denote $\lambda_{\widehat{G}}$ as the spectral gap of the self-looped graph $\widehat{G}$, that is, the least nonzero eigenvalue of the normalized Laplacian $\mathbf{L}_{sym}$. 
The node feature matrix is defined to be $\mathbf{X}\in\mathbb{R}^{v\times m}$ with $m$ dimensions of the node feature. 
For clarity, the $k$-th power of matrix $\mathbf{A}$ is given as $\mathbf{A}^{k}$, while the node embedding matrix in the $k$-th layer of the GNN is represented as $\mathbf{X}^{(k)}$. 

\subsection{Retrospection of Over-smooothing}
In this section, we briefly retrospect Over-smoothing in deep GCN, which could illustrate the background of our study.

\textbf{GCN.}~~The output of the $\ell$+$1$-th layer Graph Convolutional Networks (GCN)~\cite{kipf2016semi} is formulated as:
\begin{equation}\label{equ:gcn}
    \mathbf{X}^{(\ell+1)}=\sigma\left(\left(\widehat{\mathbf{D}}^{-\frac{1}{2}}\widehat{\mathbf{A}}\widehat{\mathbf{D}}^{-\frac{1}{2}}\right)\mathbf{X}^{(\ell)} \mathbf{W}^{(\ell+1)}\right),
\end{equation}
where $\sigma$ denotes the $\mathrm{ReLU}$ function.
The graph convolution operator $\widehat{\mathbf{D}}^{-\frac{1}{2}}\widehat{\mathbf{A}}\widehat{\mathbf{D}}^{-\frac{1}{2}}$ can be expressed in a spectral form as $\mathbf{U}\left(\mathbf{I}-\mathbf{\Lambda}\right)\mathbf{U}^{\top}$.
As we let $\ell\to\infty$, for a non-bipartite graph, the output of the $\ell$+$1$-th GCN will converge: $\mathbf{X}^{(\ell+1)} \to u_vu_v^{\top}\cdot \mathbf{X}^{(0)}\mathbf{W}^{(1)}\mathbf{W}^{(2)}...\mathbf{W}^{(\ell+1)}$.
This convergence mathematically illustrates the Laplacian smoothing process, which eventually leads to Over-smoothing.
However, we doubt that the above convergence (Over-smoothing) may not occur as early as performance degradation (see analysis in~\cref{sec:convergence rate}).
Meanwhile, the influence of the stacks of $\mathbf{W}^{(1)}\mathbf{W}^{(2)}...\mathbf{W}^{(\ell+1)}$ in the result of convergence warrants further study.

\underline{\emph{Proof of the Convergence Result.}} To simplify the formulation, the activation function $\sigma$ is omitted. Hence, the output of the $\ell$+$1$-th GCN after simplification can be expanded as:
\begin{equation}
\begin{aligned}
    \mathbf{X}^{(\ell+1)}&=\left(\widehat{\mathbf{D}}^{-\frac{1}{2}}\widehat{\mathbf{A}}\widehat{\mathbf{D}}^{-\frac{1}{2}}\right)\mathbf{X}^{(\ell)} \mathbf{W}^{(\ell+1)}\\
    &=\mathbf{U}\left(\mathbf{I}-\mathbf{\Lambda}\right)^{\ell}\mathbf{U}^{\top}\mathbf{X}^{(0)}\mathbf{W}^{(1)}\mathbf{W}^{(2)}...\mathbf{W}^{(\ell+1)}.
\end{aligned}
\end{equation}

When $\widehat{G}$ is a non-bipartite connected graph, the eigenvalues of $\widehat{\mathbf{D}}^{-\frac{1}{2}}\widehat{\mathbf{A}}\widehat{\mathbf{D}}^{-\frac{1}{2}}$ satisfy: $-1<1-a_v \leq 1-a_{v-1} \leq ... \leq 1-a_2<1-a_1=1$.
As we let $\ell\to\infty$:
\begin{equation}
\begin{aligned}
    &\lim_{\ell\to\infty}{\left(\widehat{\mathbf{D}}^{-\frac{1}{2}}\widehat{\mathbf{A}}\widehat{\mathbf{D}}^{-\frac{1}{2}}\right)}^{\ell}\mathbf{X}^{(0)}\\
    =&\lim_{\ell\to\infty}\mathbf{U}\cdot \mathrm{diag}\left({(1-a_v)}^\ell, {(1-a_{v-1})}^\ell, ... , {(1-a_1)}^\ell\right)\cdot \mathbf{U}^{\top}\mathbf{X}^{(0)}\\
    =&u_vu_v^{\top}\cdot \mathbf{X}^{(0)}.
\end{aligned}
\end{equation}

Hence, with increased layer depth of GCN, the output of the $\ell$+$1$-th GCN will converge: $\mathbf{X}^{(\ell+1)} \to u_vu_v^{\top}\cdot \mathbf{X}^{(0)}\mathbf{W}^{(1)}\mathbf{W}^{(2)}...\mathbf{W}^{(\ell+1)}$.

\begin{figure*}[t]
\vskip -0.25in
    \centering
    \subfloat[Model Performance]{\includegraphics[width=0.75\columnwidth]{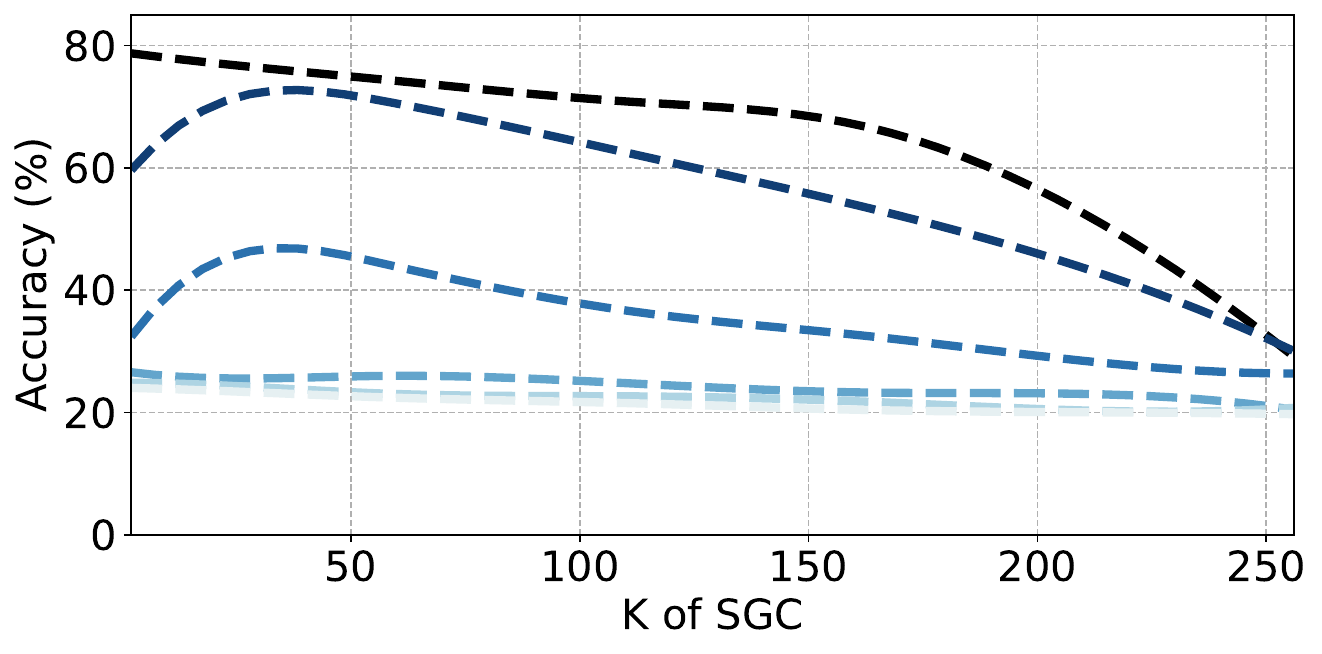}}\hspace{7mm}
    \subfloat[Dirichlet Energy $E(\mathbf{X})$]{\includegraphics[width=0.8\columnwidth]{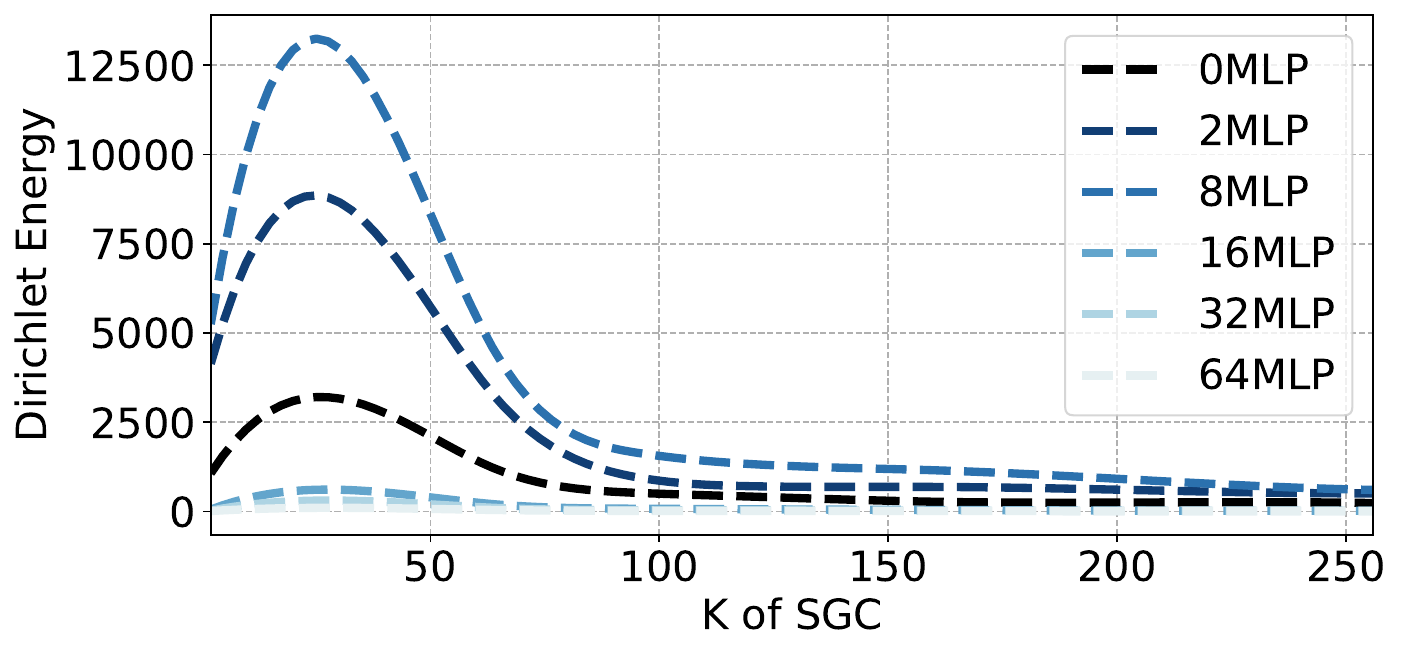}}
    \vskip -0.15in
    \caption{The test accuracy ($\%$) (a) and Dirichlet Energy (b) of node classification task of SGC ($K$= 2, 3, 4, ... ) with incremental MLP layers ($L$= 0, 2, 8, 16, 32, 64) on Cora.}
    \label{fig:SGC-MLP} 
\vskip -0.15in
\end{figure*}

\subsection{Convergence Rate of Smoothing and Spectral Gap}\label{sec:convergence rate}
The convergence rate of smoothing has been analyzed from the views of Dirichlet Energy~\cite{cai2020note,zhou2021dirichlet}, lazy random walk~\cite{chen2020simple,yang2024efficient}, $etc$. 
For Dirichlet Energy, when it converges to zero exponentially, all node embeddings are almost identical, and Over-smoothing occurs. 
For lazy random walk, it gradually converges to the stationary state and thus leads to Over-smoothing as well.

Their upper bounds of the convergence rate of smoothing are regulated by spectral gap $\lambda_{\widehat{G}}$.
For example, the upper bound of Dirichlet Energy $E(\mathbf{X})$ in~\cite{cai2020note} is given as $E((\widehat{\mathbf{D}}^{-\frac{1}{2}}\widehat{\mathbf{A}}\widehat{\mathbf{D}}^{-\frac{1}{2}})\mathbf{X})<(1-\lambda_{\widehat{G}})^2 E(\mathbf{X})$, which is relatively large due to the small $\lambda_{\widehat{G}}$ of real graphs~\cite{mohaisen2010measuring}.
However, a stringent lower bound of the convergence rate is lacking to precisely describe the occurrence of Over-smoothing. 
For instance, considering a small $\lambda_{\widehat{G}}$ of real graphs, the existing lower bound in~\cite{zhou2021dirichlet} is too broad to be effectively informative.

\textbf{Mixing Time.}~~In the analysis of Markov chain, Mixing Time measures the time required by a Markov chain for the distance to the stationary distribution $\pi$ to be small~\cite{levin2017markov}. 
The Mixing Time is defined by: $t_{mix}(\epsilon):=\min\{t:d(t)\leq\epsilon\}$, where $\epsilon>0$ and $d(t)$ denotes the maximal distance between the state at time $t$ and the final stationary state.
In~\cite{levin2017markov}, with lazy random walk transition matrix $\mathbf{P}$, its eigenvalues satisfy: $1=\lambda_1>\lambda_2\geq...\geq\lambda_v\geq-1$ and $1-\lambda_2$ is defined as the spectral gap. 
Mixing Time is proved to be bounded as:
\begin{equation}\label{equ:tmix bound}
\begin{aligned}
\frac{\lambda_2}{1-\lambda_2}\log\left(\frac{1}{2\epsilon}\right)\leq t_{mix}(\epsilon)\leq\frac{1}{1-\lambda_2}\log\left(\frac{1}{\epsilon\pi_{min}}\right).
\end{aligned}
\end{equation}
\cite{wang2019improving} proves that GCN with simple residual connection simulates a lazy random walk with transition matrix $\mathbf{P}=\frac{\mathbf{I}+\widehat{\mathbf{D}}^{-\frac{1}{2}}\widehat{\mathbf{A}}\widehat{\mathbf{D}}^{-\frac{1}{2}}}{2}$, which gradually converges to the stationary state and thus leads to Over-smoothing. 
Since the lazy random walk is an ergodic Markov chain, the Mixing Time of GCN with a simple residual connection will also be bounded as (\ref{equ:tmix bound}), and it could be a suitable measurement to describe the convergence rate of smoothing.
Aligning with the normalized Laplacian matrix $\mathbf{L}_{sym}$, it is straightforward to have $\lambda_i=1-\frac{a_i}{2}$ and ${a_2}={\lambda_{\widehat{G}}}$.
Consequently, we prove that Mixing time is bounded as follows:
\begin{equation}\label{equ:new tmix bound}
\begin{aligned}
\frac{2-{\lambda_{\widehat{G}}}{}}{{\lambda_{\widehat{G}}}{}}\log\left(\frac{1}{2\epsilon}\right)\leq t_{mix}(\epsilon)\leq
\frac{2}{\lambda_{\widehat{G}}}\log\left(\frac{1}{\epsilon\pi_{min}}\right).
\end{aligned}
\end{equation}

In practice, real graphs like social networks are usually sparse, with extremely small $\lambda_{\widehat{G}}$~\cite{mohaisen2010measuring}, which leads to large lower and upper bounds for Mixing Times. 
Therefore, from the perspective of Mixing Time, {\emph{we can conclude that the convergence rate of smoothing on most real graphs is considerably slower than that of performance degradation}} (at around 8 layers for GCN~\cite{kipf2016semi}).

\section{Empirical Analysis on Deep GNNs}
In this section, we analyze the dominant problem of deep GNNs and investigate where these so-called deep models towards solving Over-smoothing actually work from an empirical perspective.

\subsection{Over-smoothing Versus Trainability}\label{sec:empirical analysis}
We investigate the dominant problem of deep GNNs through SGC-MLP experiments and Dirichlet Energy~\cite{cai2020note} analysis.

\textbf{SGC-MLP.}~~~To be specific, we evaluate the node classification performance of vanilla SGC~\cite{wu2019simplifying} ($K$= 2, 3, 4, ... ) with incremental MLP layers ($L$= 0, 2, 8, 16, 32, 64), which inherently decouple the graph propagation process and the training process.
The output of SGC-MLP is formulated as:
\begin{equation}
\begin{aligned}
    \mathbf{Z}=
    \underbrace{\left(\left(\widehat{\mathbf{D}}^{-\frac{1}{2}}\widehat{\mathbf{A}}\widehat{\mathbf{D}}^{-\frac{1}{2}}\right)^{K}\mathbf{X}^{(0)}\mathbf{W}\right)}_{\text{vanilla SGC}}\underbrace{\mathbf{W}^{(1)}\mathbf{W}^{(2)}\mathbf{W}^{(3)}...\mathbf{W}^{(L)}}_{\text{incremental MLP Layers}}
\end{aligned}
\end{equation}

\textbf{Dirichlet Energy.}~~~Additionally, we use Dirichlet Energy $E(\mathbf{X})$~\cite{cai2020note} to measure the smoothness of SGC-MLP.
A small value of $E(\mathbf{X})$ reflects the high similarity of the node embeddings, which is highly related to the Over-smoothing~\cite{zhou2021dirichlet}.
The embedding vector of the $i$-th node is given as $\mathbf{X}_i$.
Dirichlet Energy $E(\mathbf{X})$ is defined as:
\begin{equation}
\begin{aligned}
E(\mathbf{X})=\frac{1}{v}\sum_{i\in{V}}\sum_{j\in\mathcal{N}_i}\|\mathbf{X}_i-\mathbf{X}_j\|_{2}^{2}.
\end{aligned}
\end{equation}

{\textbf{Over-smoothing is not the dominant factor that leads to the performance drop of deep GNNs.}}
According to the characteristics of Over-smoothing, even if $L$ is small, the model performance and Dirichlet Energy will decrease significantly as $K$ increases slightly, which is in contradiction with the results reflected in~\cref{fig:SGC-MLP}.
Firstly, from the perspective of model performance in~\cref{fig:SGC-MLP}(a), in the case of no or merely 2 MLP layers, SGC-MLP with deep graph propagations does not suffer greatly from the smoothing process until the propagation order $K$ exceeds about 120.
However, the introduction of around 8 MLP layers leads to a significant drop in model performance. 
In~\cref{fig:SGC-MLP}(b), we focus on the periods of sharp drops in Dirichlet Energy, which indicate that the degree of smoothness is increasing.
For SGC-MLP with shallow MLP layers ($L$ < 8), Dirichlet Energy begins to plummet when $K$ is around 30, which occurs much later than the performance degradation in GCN (at around 8 layers~\cite{kipf2016semi}).
Meanwhile, when the number of MLP layers increases to 16, 32, or 64, the level of smoothness of SGC-MLP is always high, even if the graph propagation order $K$ is less than 10.
This indicates that deep MLP greatly aggravates the performance degradation of the model, and then leads to the superficial problem, Over-smoothing.
Actually, it is the trainability issue that mainly leads to the performance drop.

\begin{figure*}[t]
    \centering
    \vskip -0.15in
    \includegraphics[width=1.7\columnwidth]{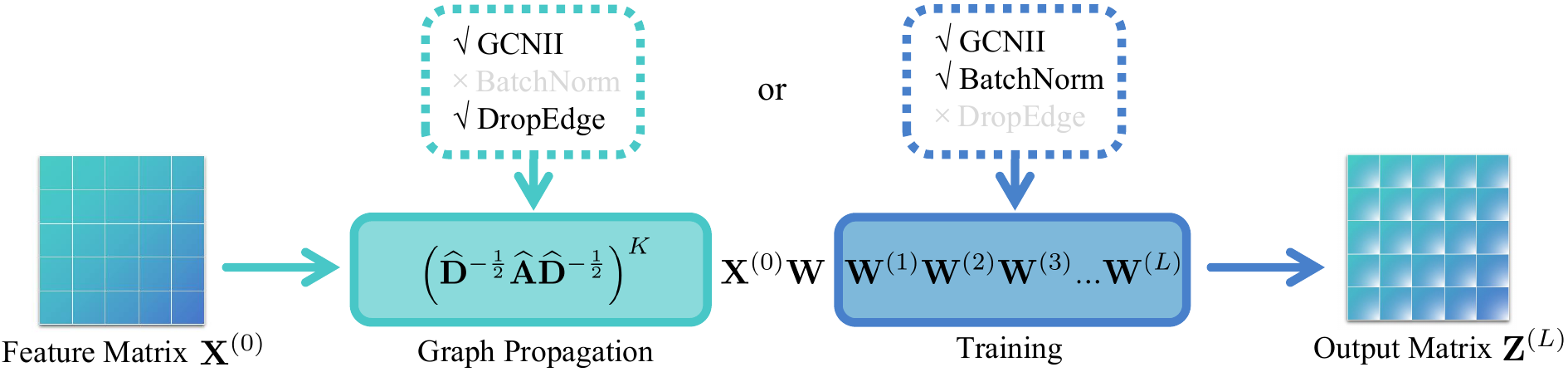}\\
    \vspace{1mm}
    \includegraphics[width=1.0\columnwidth]{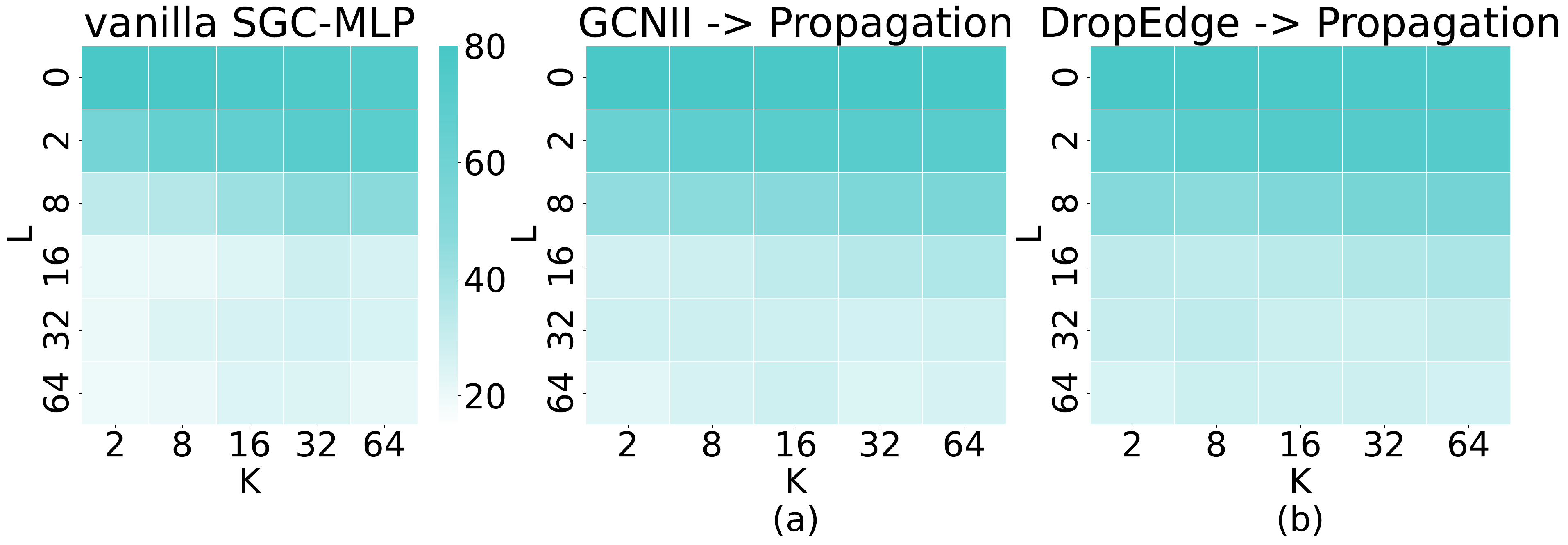}\hspace{3mm}
    \includegraphics[width=1.0\columnwidth]{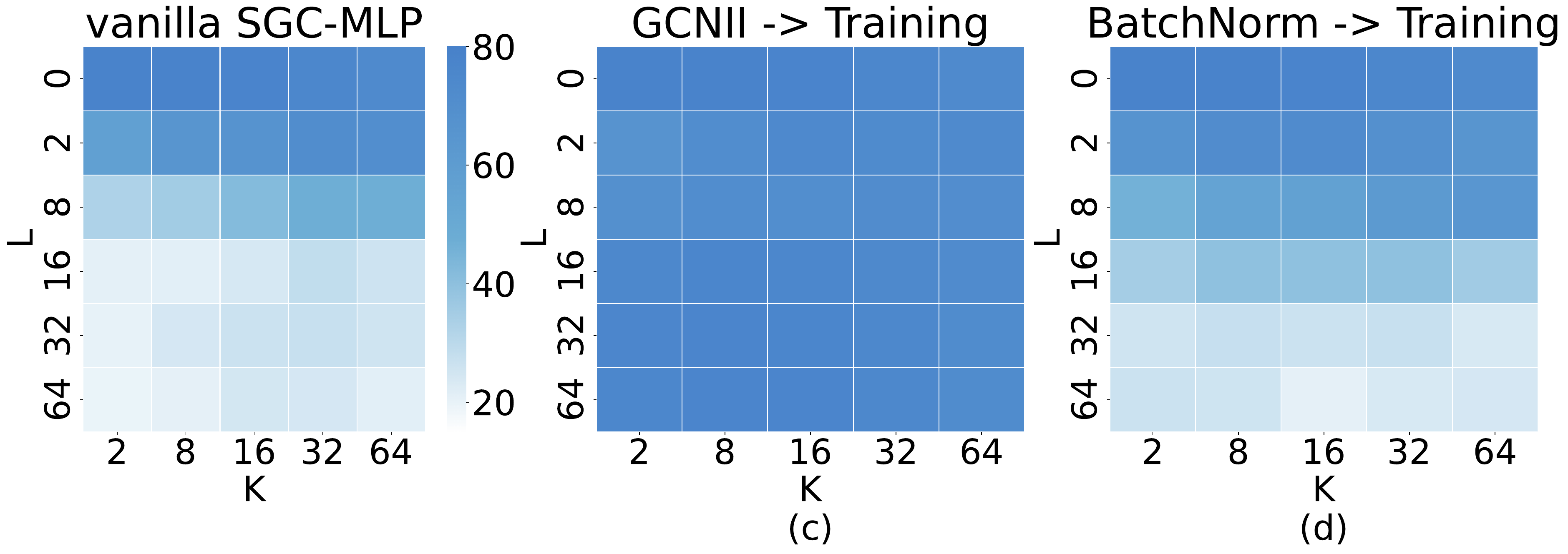}
    \vskip -0.1in
    \caption{Illustration of decoupled experiments (above). The decoupled experiments investigate the actual effectiveness of the tricks of GCNII, Batch Normalization, and DropEdge by adding them separately to the graph propagation process or the training process. Results of decoupled experiments (below). The contrast of color shades with vanilla SGC-MLP on the left reflects a decrease or increase in the accuracy ($\%$) of node classification tasks on Cora after using various tricks in each process.}
    \label{fig:SGC-MLP2}  
\vskip -0.1in
\end{figure*}

\begin{table}[t]
    \caption{The formulas of the graph propagation and the training process after applying the tricks of GCNII, Batch Normalization, and DropEdge to the \textbf{former} or the latter. $\alpha$ and $\beta$ are parameters introduced by the techniques of GCNII. The Batch Normalization layer is abbreviated by $BN$. $\widehat{\mathbf{A}}_{drop}^{(K-1)}$ denotes the new adjacency matrix augmented by DropEdge at the $K$-1-th layer. $Drop$ denotes the drop operation like DropOut. $\mathbf{Z}^{(0)}=\mathbf{X}^{(K)}\mathbf{W}$. }
    \vskip -0.1in
    \label{tab:formula}
    \centering
    \resizebox{\columnwidth}{!}{\begin{tabular}{l|c|c}
    \toprule
    &  Graph Propagation&  Training\\
    \midrule
    Vanilla&   $\mathbf{X}^{(K)}= \left(\widehat{\mathbf{D}}^{-\frac{1}{2}}\widehat{\mathbf{A}}\widehat{\mathbf{D}}^{-\frac{1}{2}}\right)\mathbf{X}^{(K-1)} $& \multirow{5}{*}{$\mathbf{Z}^{(L)}=\mathbf{Z}^{(L-1)}\mathbf{W}^{(L-1)}$}\\
    GCNII&  $\mathbf{X}^{(K)}=\left(1-\alpha^{\left(K-1\right)}\right) \left(\widehat{\mathbf{D}}^{-\frac{1}{2}}\widehat{\mathbf{A}}\widehat{\mathbf{D}}^{-\frac{1}{2}}\right)\mathbf{X}^{(K-1)}+\alpha^{(K-1)}\mathbf{X}^{(0)}$&  \\
    BatchNorm& $\mathbf{X}^{(K)}= BN^{\left(K-1\right)}\left(\left(\widehat{\mathbf{D}}^{-\frac{1}{2}}\widehat{\mathbf{A}}\widehat{\mathbf{D}}^{-\frac{1}{2}}\right)\mathbf{X}^{(K-1)}\right) $& \\
    DropEdge& $\mathbf{X}^{(K)}= \left(\widehat{\mathbf{D}}^{-\frac{1}{2}}\widehat{\mathbf{A}}_{drop}^{(K-1)}\widehat{\mathbf{D}}^{-\frac{1}{2}}\right)\mathbf{X}^{(K-1)}$& \\
 \bottomrule
    \end{tabular}}
    \begin{tablenotes}
        \footnotesize
        \item Note: Tricks are applied to the \textbf{Graph Propagation process}.
    \end{tablenotes}
\vskip -0.15in
\end{table}

\begin{table}[t]
    \label{tab:formula_train}
    \centering
    \resizebox{\columnwidth}{!}{
    \begin{tabular}{l|c|c}
    \toprule
    &  Graph Propagation&  Training\\
    \midrule
    Vanilla&   \multirow{6}{*}{$\mathbf{X}^{(K)}= \left(\widehat{\mathbf{D}}^{-\frac{1}{2}}\widehat{\mathbf{A}}\widehat{\mathbf{D}}^{-\frac{1}{2}}\right)\mathbf{X}^{(K-1)}$}& $\mathbf{Z}^{(L)}=\mathbf{Z}^{(L-1)}\mathbf{W}^{(L-1)}$\\
    GCNII&  &  $\makecell[c]{\mathbf{Z}^{(L)}=\left(1-\alpha^{(L-1)}\right)(\mathbf{Z}^{(L-1)} ((1-\beta^{(L-1)})\mathbf{I}+\\ \beta^{(L-1)}\mathbf{W}^{(L-1)}))+\alpha^{(L-1)}\mathbf{Z}^{(0)}}$\\
    BatchNorm& & $\mathbf{Z}^{(L)}=BN^{(L-1)}\left(\mathbf{Z}^{(L-1)}\mathbf{W}^{(L-1)}\right)$\\
    DropEdge& & $\mathbf{Z}^{(L)}=Drop^{(L-1)}\left(\mathbf{Z}^{(L-1)}\mathbf{W}^{(L-1)}\right)$\\
    \bottomrule
    \end{tabular}}
    \begin{tablenotes}
        \footnotesize
        \item Note: Tricks are applied to the \textbf{Training process}.
    \end{tablenotes}
\vskip -0.1in
\end{table}

\subsection{Revisit GNNs via Decoupled Experiments}\label{sec:revisit exp}
The analysis in~\cref{sec:empirical analysis} indicates that Over-smoothing is not the dominant problem affecting the model performance. 
However, many deep GNNs in the past have claimed that the model performance is improved by effectively addressing Over-smoothing, where \textbf{attribution errors} may exist. 
Therefore, in this section, we further analyze whether these deep GNNs alleviate Oversmoothing or solve other issues, like the trainability problem. 
To be specific, we revisit GCNII~\cite{chen2020simple}, Batch Normalization~\cite{ioffe2015batch}, and DropEdge~\cite{rong2019dropedge}, three representative methods of Residual Connection, Normalization, and Drop Operation.
Then, based on the experimental settings of SGC-MLP in~\cref{sec:empirical analysis}, we investigate the actual effectiveness of the tricks of each model by adding them separately to the graph propagation process or the training process. The details and results of decoupled experiments are illustrated in~\cref{fig:SGC-MLP2}.
The corresponding formula structures are shown in~\cref{tab:formula}.

\begin{table*}[t]
\vskip -0.05in
\caption{Summary of semi-supervised node classification accuracy ($\%$) of GCN(Batch Normalization), GCN(DropEdge), and GCNII with various depths. Batch Normalization and DropEdge are abbreviated by BN and DE, respectively.}
\vskip -0.1in
\label{tab:acc-bn-de-ii-new}

\begin{center}
\begin{small}
\begin{tabular}{l|ccccc|ccccc|ccccc}
\toprule
\multirow{1}{*}{Dataset} &  \multicolumn{5}{c|}{Cora} &  \multicolumn{5}{c|}{Citeseer} &  \multicolumn{5}{c}{Pubmed}  \\
Layers&  2 &8& 16&  32 &64& 2 &8& 16&  32 &64& 2 &8& 16& 32 &64\\
\midrule
GCN&  \textbf{81.1} &69.5& 64.9&  60.3&28.7& \textbf{70.8} &30.2&  18.3&  25.0 &20.0 &\textbf{79.0} &61.2&  40.9& 22.4&35.3\\

GCN (Batch Normalization)&  \textbf{77.6}&74.5& 72.3& 66.3&55.9&\textbf{57.2}&51.6&47.1& 46.8&43.3&\textbf{76.4}&74.9&73.2&71.5&63.3\\

GCN (DropEdge)&  \textbf{82.8} &75.8& 75.7&  62.5&49.5&\textbf{72.3} &61.4& 57.2&  41.6&34.4&\textbf{79.6} &78.1& 78.5& 77.0  &61.5\\

GCNII&  82.2 &84.2& 84.6&  85.4&\textbf{85.5}&68.2 &70.6& 72.9&   \textbf{73.4}&73.4&78.2 &79.3& \textbf{80.2}&  79.8 &79.7\\
\bottomrule
\end{tabular}
\end{small}
\end{center}
\vskip -0.1in
\end{table*}
\textbf{GCNII effectively addresses the dominant problem of deep GNNs, the trainability issue.}
For GCNII, we conduct experiments by incorporating \emph{Initial residual} and \emph{Identity mapping}, two techniques of GCNII, into the decoupled graph propagation process and training process, respectively.
The results are witnessed in~\cref{fig:SGC-MLP2}(a) and (c), which illustrate that two techniques of GCNII are effective in the graph propagation process and the training process.
However, the improvement in the graph propagation process is limited.
Especially when the number of MLP layers is further increased, the model performance will significantly decrease as well.
In contrast, the improvement in the training process is notable.
This explains the superior performance of deep GCNII (shown in~\cref{tab:acc-bn-de-ii-new}); that is, GCNII effectively addresses the dominant problem of deep GNNs by enhancing trainability rather than mitigating Over-smoothing by improving the graph propagation process.

We then select \emph{Batch Normalization} and \emph{DropEdge} in the category of Normalization and Drop Operation, respectively. 
However, the graph propagation process naturally lacks learnable parameters, which is inconsistent with the role of Batch Normalization layers.
Similarly, for DropEdge, no new graph information is used as input in the process of stacking MLPs. 
Consequently, DropEdge can merely be applied in the graph propagation process and is not suitable for use in deep MLPs.
Therefore, in essence, Batch Normalization cannot directly optimize the graph propagation process, and DropEdge cannot directly improve the training process.
The results of the corresponding experiments are given in~\cref{app:decouple}, which is merely for experimental completeness.

\textbf{Batch Normalization solves the training challenges to a certain extent. DropEdge works on the side issue, Over-smoothing.}
The results of the decoupled experiments on Batch Normalization are given in~\cref{fig:SGC-MLP2}(d).
For Batch Normalization, the improvement in the training process is visible in the scenario of shallow MLPs, but its effect diminishes when MLP goes deeper. 
Combined with the performance drop of deep GCN(BatchNorm) in the coupled case in~\cref{tab:acc-bn-de-ii-new}, it indicates that Batch Normalization aids deep GNNs primarily by addressing the training challenges of deep MLPs to a certain extent, but is not significantly effective.
The results of the decoupled experiments on DropEdge are given in~\cref{fig:SGC-MLP2}(b).
For DropEdge, the mitigation of the smoothing process is visible.
Although DropEdge has a certain easing effect on Over-smoothing, the performance of GCN(DropEdge) will still significantly decrease with increased layer depth (shown in~\cref{tab:acc-bn-de-ii-new}) since DropEdge does not directly work on the dominant problem of deep GNNs.
The underlying mechanism of these techniques in improving model performance will be further analyzed in~\cref{sec:revisit gradient}.

\section{Theoretical Analysis on Deep GNNs}
In this section, we further investigate the gradient of deep GNNs.
We aim to revisit the dominant problem of deep GNNs and identify the actual effectiveness of the models that supposedly claim to solve Over-smoothing from a novel theoretical perspective.

\subsection{Analysis on Gradient Upper Bound}\label{sec:theoretical analysis}
Based on the empirical analysis in~\cref{sec:empirical analysis}, we identify that the primary reason for the poor performance of deep GNNs is not Over-smoothing but the difficult training problem of deep MLPs. 
Actually, the issue of poor trainability (deeper neural networks are more difficult to train) is a common problem in Deep Neural Networks (DeepNN)~\cite{he2016deep,arora2018convergence}. 
However, previous research on deep GNN was biased by the overstated Over-smoothing problem and there has been limited focus on theoretically elucidating the respective impacts of Over-smoothing and the training problem. 
For instance,~\citep{zhang2022model} arrives at their conclusions solely on empirical analysis, which is not sufficiently credible because of the lack of theoretical analysis;~\citep{jaiswal2022old} rewires vanilla GCN to solve the training problem with an assumption that improving gradient flow can enhance the performance of deep GCN, which lacks the analysis towards the impact of Over-smoothing and a clear motivation for analyzing the gradient;~\citep{cong2021provable} focuses on exploring the generalization capability of GCNs with the premise that the training problem of deep GCN can be solved effectively, while the performance of several deep GCN-like models during training is actually poor and the experimental setting of model depth in~\cite{cong2021provable} is shallow ($\leq 10$).

Hence, in this section, we get insight into this problem in deep GNNs by \textbf{gradient analysis} since the gradient updating process is firmly associated with the main problem of deep GNNs, trainability.
The gradient analysis not only contributes to a deep understanding of trainability but also fills a theoretical gap in previous works.

Following the simplification in~\citep{rusch2022graph}, we assume that the dimension $m$ of node feature embedding is 1 for simplicity of exposition without losing computational accuracy.     
The loss function of the deep GNNs is defined as $\mathbf{L}(\mathbf{W})=\frac{1}{2v}\sum_{i\in\mathcal{V}}\|\mathbf{X}_{i}^{(N)}-{\mathbf{Y}}_{i}\|^{2}_2$, which computes a $p$ norm of the gap between the final feature embedding $\mathbf{X}^{(N)}\in\mathbb{R}^{v}$ after $N$ layers and the corresponding target truth $\mathbf{Y}\in\mathbb{R}^{v}$, where we generally choose $p=2$. 
The objective of the learning task is to make the output of GNN closer to the target truth. 
The model is trained by minimizing the $\ell_2$ loss function with a stochastic gradient descent (SGD) procedure~\cite{zinkevich2010parallelized}. 
We consider a deep GNN with $N$ layers (depth of $N$) and $\mathbf{W}^{(\ell)}$ is the learnable weight parameters of the $\ell$-th layer, where $1\leq \ell \leq N$. 
During the training, the minimization of $\mathbf{L}(\mathbf{W})$ is conducted by iteratively applying the following updates:
\begin{equation}\label{equ:W update}
    \mathbf{W}^{(\ell)}(t)-\eta\frac{\partial \mathbf{L}(\mathbf{W})}{\partial \mathbf{W}^{(\ell)}(t)}\rightarrow \mathbf{W}^{(\ell)}(t+1), t=0,1,2,...,
\end{equation}
where $\eta>0$ is the configurable learning rate. 

Formally, the specific calculation process of a node-wise $\partial \mathbf{L}(\mathbf{W})$ according to the chain rule is given as:
\begin{equation}\label{equ:gradient}
    \frac{\partial\mathbf{L}(\mathbf{W})}{\partial{\mathbf{W}}^{(\ell)}_k}=\frac{\partial\mathbf{L}(\mathbf{W})}{\partial{\mathbf{X}}^{(N)}}\cdot\frac{\partial{\mathbf{X}}^{(N)}}{\partial{\mathbf{X}^{(\ell)}}}\cdot\frac{\partial{\mathbf{X}}^{(\ell)}}{\partial{\mathbf{W}^{(\ell)}_k}},
\end{equation}
where $\mathbf{W}^{(\ell)}_k$ is the weight parameter of the $k$-th node in the $\ell$-th layer of GNNs. 

For ease of computation, the~\cref{equ:gradient} can be written as:
\begin{equation}\label{equ:gradient calculation}
\begin{aligned}
     \frac{\partial\mathbf{L}(\mathbf{W})}{\partial{\mathbf{W}}^{(\ell)}_k}=\frac{\partial\mathbf{L}(\mathbf{W})}{\partial{\mathbf{X}}^{(N)}}\cdot
    \prod_{n=\ell+1}^{N}\frac{\partial\mathbf{X}^{(n)}}{\partial{\mathbf{X}}^{(n-1)}}\cdot\frac{\partial{\mathbf{X}}^{(\ell)}}{\partial{\mathbf{W}^{(\ell)}_k}}.
\end{aligned}
\end{equation}

\begin{table*}[t]
\vskip -0.05in
\caption{The node-wise gradient upper bounds of GCN, GCNII, GCN(Batch Normalization), and GCN(DropEdge) (first 4 rows), and the node-wise gradient upper bounds of ResGCN and $\omega$GCN (last 2 rows).} 
\label{tab:GNN_gradient bound}
\begin{center}
\begin{small}
\vskip -0.1in
\resizebox{1.52\columnwidth}{!}{
\begin{tabular}{lc}
\toprule
 Method& Node-wise Gradient Upper Bound $\left|\frac{\partial\mathbf{L}(\mathbf{W})}{\partial{\mathbf{W}}^{(\ell)}_k}\right|$\\
\midrule
 GCN& $\frac{\Gamma}{v}\cdot\left(\Delta\right)^{N-\ell+1}\cdot\prod_{n=\ell+1}^{N}\left\|\mathbf{W}^{(n)}\right\|_\infty\cdot\left\|\mathbf{X}^{(\ell-1)}\right\|_\infty$\\
 
 GCNII& ${\frac{\Gamma}{v}\cdot\left(\left(1-\alpha\right)\Delta\right)^{N-\ell}\cdot\prod_{n=\ell+1}^{N}\left(\left(1-\beta\right)+\beta\left\|\mathbf{W}^{(n)}\right\|_\infty\right)\cdot\left(\beta\left\|\mathbf{X}^{(\ell-1)}\right\|_\infty+\alpha\beta\left\|\mathbf{X}^{(0)}\right\|_\infty\right)}$\\

GCN(Batch Normalization)& $\frac{\Gamma}{v}\cdot\left(\Delta\right)^{N-\ell+1}\cdot\prod_{n=\ell+1}^{N}\frac{\gamma^{(n)}}{\sigma^{(n)}_{BN}}\left\|\mathbf{W}^{(n)}\right\|_\infty\cdot\frac{\gamma^{(\ell)}}{\sigma^{(\ell)}_{BN}}\left\|\mathbf{X}^{(\ell-1)}\right\|_\infty$\\

  GCN(DropEdge)& $\frac{\Gamma}{v}\cdot\prod_{n=\ell}^{N}\left(\Tilde{\Delta}^{(n)}\right)\cdot\prod_{n=\ell+1}^{N}\left\|\mathbf{W}^{(n)}\right\|_\infty\cdot\left\|\mathbf{X}^{(\ell-1)}\right\|_\infty$\\
\midrule
  ResGCN& $\frac{\Gamma}{v}\cdot\prod_{n=\ell+1}^{N}\left(\Delta\left\|\mathbf{W}^{(n)}\right\|_\infty+1\right)\cdot\Delta\left\|\mathbf{X}^{(\ell-1)}\right\|_\infty$\\

$\omega$GCN & $\frac{\Gamma}{v}\cdot\left(\omega\Delta+1-\omega\right)^{N-\ell+1}\cdot\prod_{n=\ell+1}^{N}\left\|\mathbf{W}^{(n)}\right\|_\infty\cdot\left\|\mathbf{X}^{(\ell-1)}\right\|_\infty$\\
\bottomrule
\end{tabular}
}
\begin{tablenotes}
    \footnotesize
    \item Note: $\Gamma$ denotes $\left(\max_{1\leq i\leq v}|\mathbf{X}_i^{(N)}|+\max_{1\leq i\leq v}|{\mathbf{Y}}_i|\right)$. $\Delta$ denotes $\max_{1\leq i\leq v,j\in\mathcal{N}_i}\frac{1}{\sqrt{d_id_j}}$. $\frac{\gamma^{(\ell)}}{\sigma^{(\ell)}_{BN}}$ denotes the gradient of the $\ell$-th Batch Normalization layer. $\Tilde{\Delta}^{(\ell)}$ denotes $\max_{1\leq i\leq v,j\in{\mathcal{N}}_i^{(\ell)}}\frac{1}{\sqrt{{d}_i^{(\ell)}{d}_j^{(\ell)}}}$, which reflects the input graph structure information of the $\ell$-th layer after DropEdge. $\omega$ is a learnable parameter in $\omega$GCN.
\end{tablenotes}
\end{small}
\end{center}
\vskip -0.2in
\end{table*}

In order to get insight into the node-wise gradient of GCN, the node-wise form of deep GCN is defined as:
\begin{equation}\label{equ:node-wise GCN}
{\mathbf{X}}_i^{(\ell+1)}=\sigma\left(\frac{{\mathbf{W}}_i^{(\ell+1)}}{d_i}{\mathbf{X}}_i^{(\ell)}+\sum_{j\in\mathcal{N}_i}\frac{{\mathbf{W}}_j^{(\ell+1)}}{\sqrt{d_id_j}}{\mathbf{X}}_j^{(\ell)}\right).
\end{equation}

\begin{proposition}\label{pro:GCN graident bound}
The node-wise gradient of GCN with regards to any learnable weight parameter ${\mathbf{W}}^{(\ell)}_k$, for $1\leq k\leq v$, $1\leq\ell\leq N$ is bounded as (proved in the~\cref{app:proof1}):
\begin{equation}
\begin{aligned}
    &\left|\frac{\partial\mathbf{L}(\mathbf{W})}{\partial{\mathbf{W}}^{(\ell)}_k}\right|
    \leq\frac{1}{v}\left(\max_{1\leq i\leq v}|{\mathbf{X}}_i^{(N)}|+\max_{1\leq i\leq v}|{\mathbf{Y}}_i|\right)\cdot\\
    &\underbrace{\left(\max_{1\leq i\leq v,j\in\mathcal{N}_i}\frac{1}{\sqrt{d_id_j}}\right)^{N-\ell+1}}_{\text{smoothing term}}\underbrace{\prod_{n=\ell+1}^{N}\left\|\mathbf{W}^{(n)}\right\|_\infty}_{\text{training term}}\cdot\left\|\mathbf{X}^{(\ell-1)}\right\|_\infty.
\end{aligned}
\end{equation}
\end{proposition}

It is observed that the graph propagation process will eventually be characterized by the \emph{smoothing term}, firmly correlated with the degree information in the node-wise gradient upper bound, while stacks of MLPs will be expressed as the \emph{training term}, the product of the norms of multiple weights.
Both of them have influence on the gradient upper bound, but the extent of their impacts varies.

\textbf{The training term has a greater influence than the smoothing term.}
The key distinction between them is that the smoothing term corresponding to the degree information is a fixed value determined by the density of the graph. 
Its impact on the gradient upper bound remains constant at a certain model depth, but it will converge to zero when the model is extremely deep, potentially resulting in the gradient vanishing. 
Combined with our analysis on Mixing Time in~\cref{sec:convergence rate}, the convergence rate of the smoothing term is also tempered by the sparsity of the most real graphs~\cite{mohaisen2010measuring}, with $\max_{1\leq i\leq v,j\in\mathcal{N}_i}\frac{1}{\sqrt{d_id_j}}$ closing to 1.
Hence, the impact of the smoothing process is limited, unless the model is extremely deep or considers dense graphs.
However, the influence of the deep MLPs, the training term, is contingent on their learning results. 
Should the ideal weights not be effectively learned, gradient fluctuations could become pronounced, even in the scenario of shallow GNN, leading to early performance degradation. 
This highlights the trainability of MLPs as a pivotal problem in GNNs.

Therefore, from the perspective of the gradient upper bound, the graph propagation process does impact the gradient, but \emph{Over-smoothing is an influential problem only in cases of extremely deep GNNs or exceedingly dense graphs}. 
In contrast, the training results of deep MLPs exert a more substantial influence on the gradient at an earlier stage, which suggests that \emph{the trainability problem is the dominant factor for deep GNNs}. 
This conclusion is consistent with the observations in our empirical analysis in~\cref{sec:empirical analysis}.

\subsection{Revisit GNNs via Gradient Analysis}\label{sec:revisit gradient}
In~\cref{sec:revisit exp}, we empirically revisit three representative methods to examine whether these deep GNNs mitigate Over-smoothing or address the challenging training problems of deep MLPs via decoupled experiments. 
In this section, we will delve deeper into this issue via detailed gradient analysis in a theoretical way. 

Aligning with the settings in~\cref{sec:revisit exp}, we choose GCNII, GCN(Batch Normalization), and GCN(DropEdge) as the models for exploration. 
Since we choose GCN-like models as the backbone, the graph propagation process and the training process are coupled.
Specifically, the theoretical analysis commences with the derivation of node-wise gradient upper bounds for these deep GNN models, as presented in the first 4 rows in~\cref{tab:GNN_gradient bound}.
For details of derivation, please refer to~\cref{extral_proofs}. 
Subsequently, based on the analysis in~\cref{sec:theoretical analysis}, we can explore where these GNN models work by discerning the difference between the smoothing term and the training term in the gradient upper bound for each GNN.

\textbf{GCNII effectively regulates the training term.}
For GCNII, the smoothing term and the training term are modulated by parameters $\alpha$ and $\beta$, introduced by \emph{Initial residual} and \emph{Identity mapping}, two techniques of GCNII. 
Hence, {GCNII controls the graph propagation process and the training process} by constraining the smoothing term and the training term to a properly small level to keep a stable gradient during training, which is further analyzed in~\cref{sec:deep insight}.
This dual regulation effectively enhances the trainability, the dominant problem of deep GNNs, offering an explanation for the good performance of deep GCNII.

\textbf{Batch Normalization merely controls the training term to a certain extent.}
For GCN(Batch Normalization), the Batch Normalization layer between each GCN layer introduces a learnable term $\frac{\gamma^{(\ell)}}{\sigma^{(\ell)}_{BN}}$, affecting the norm of the weight parameters due to the gradient of Batch Normalization layers~\cite{ioffe2015batch}.
Notably, the smoothing term remains unaffected, indicating that Batch Normalization does not directly influence the smoothing process from a theoretical view.
Consequently, the improved performance of deep GNNs with Batch Normalization is attributed to the fact that {it mitigates the dominant problem, the trainability issues}. 
However, the effectiveness of each Batch Normalization layer is still contingent on the learning result of $\frac{\gamma^{(\ell)}}{\sigma^{(\ell)}_{BN}}$, so the performance of deep GNNs declines for poor training.

\textbf{DropEdge essentially modifies the smoothing term.}
For GCN(DropEdge), in the scenario of layer-wise DropEdge~\cite{rong2019dropedge}, the input graph structure information of each GCN layer is augmented in a random way.
This results in more intricate degree information, indicating that {DropEdge substantially modifies the smoothing term and works on the graph propagation process}. 
Meanwhile, DropEdge indirectly influences the learning process of the weights to a certain extent because the gradient upper bound is affected by the smoothing term as well. 
Nevertheless, DropEdge does not essentially address the dominant issue of deep GNNs, which makes it unable to prevent performance degradation in deeper models.

\begin{figure}[b]
    \centering
    \vskip -0.15in
    \includegraphics[width=0.75\columnwidth]{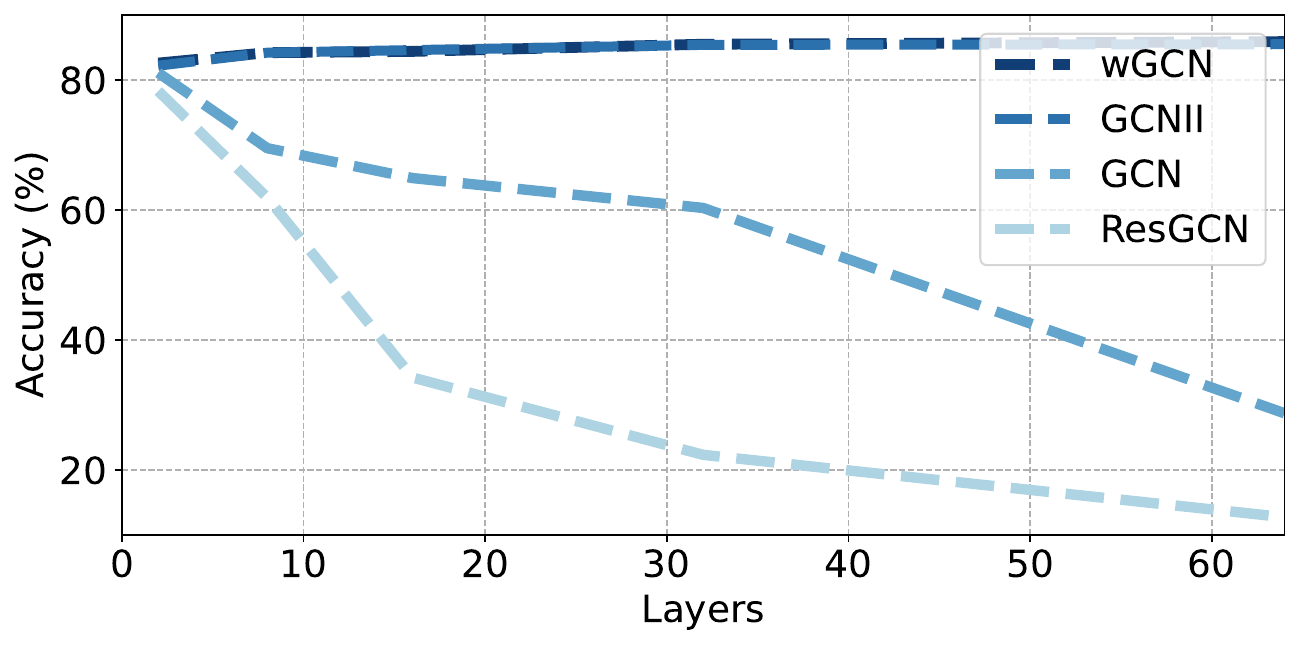}
    \vskip -0.13in
    \caption{The test accuracy ($\%$) of ResGCN, GCN, GCNII, and $\omega$GCN with increased layer depth on Cora.}
    \label{fig:ACC}
\end{figure}

\begin{figure*}[t]
    \centering
    \vskip -0.15in
    \includegraphics[width=0.67\columnwidth]{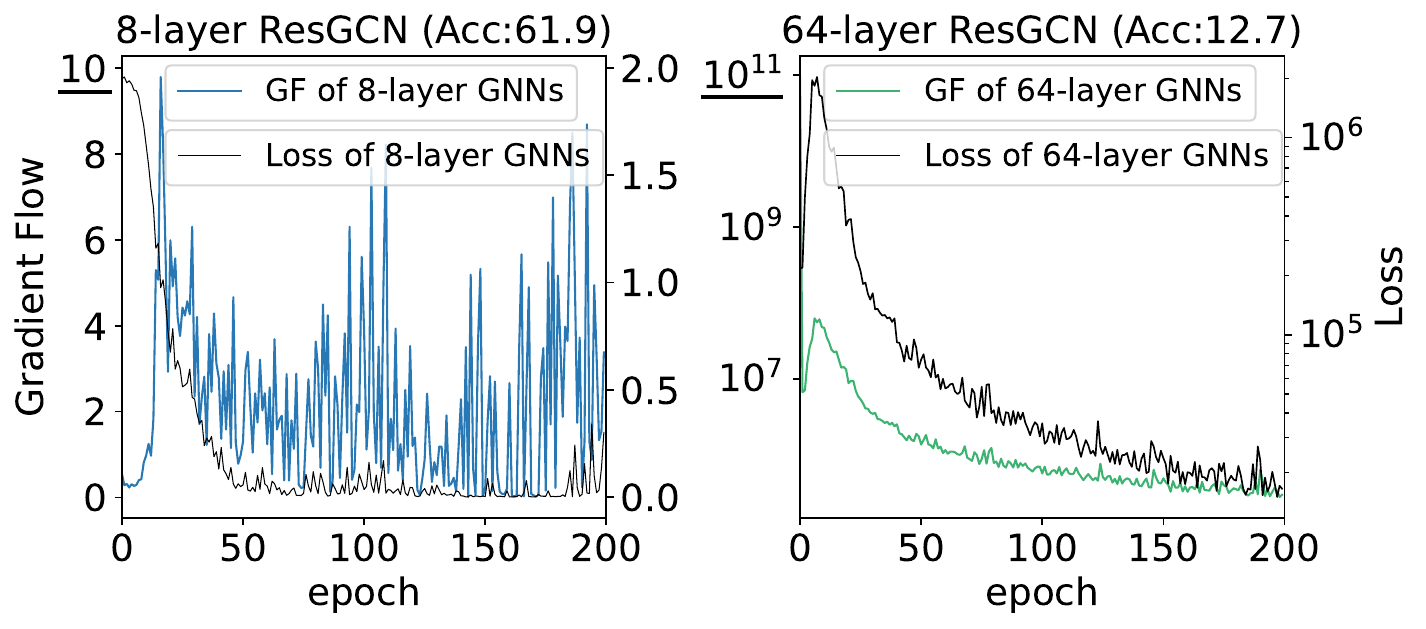}\hspace{3mm}
    \includegraphics[width=0.67\columnwidth]{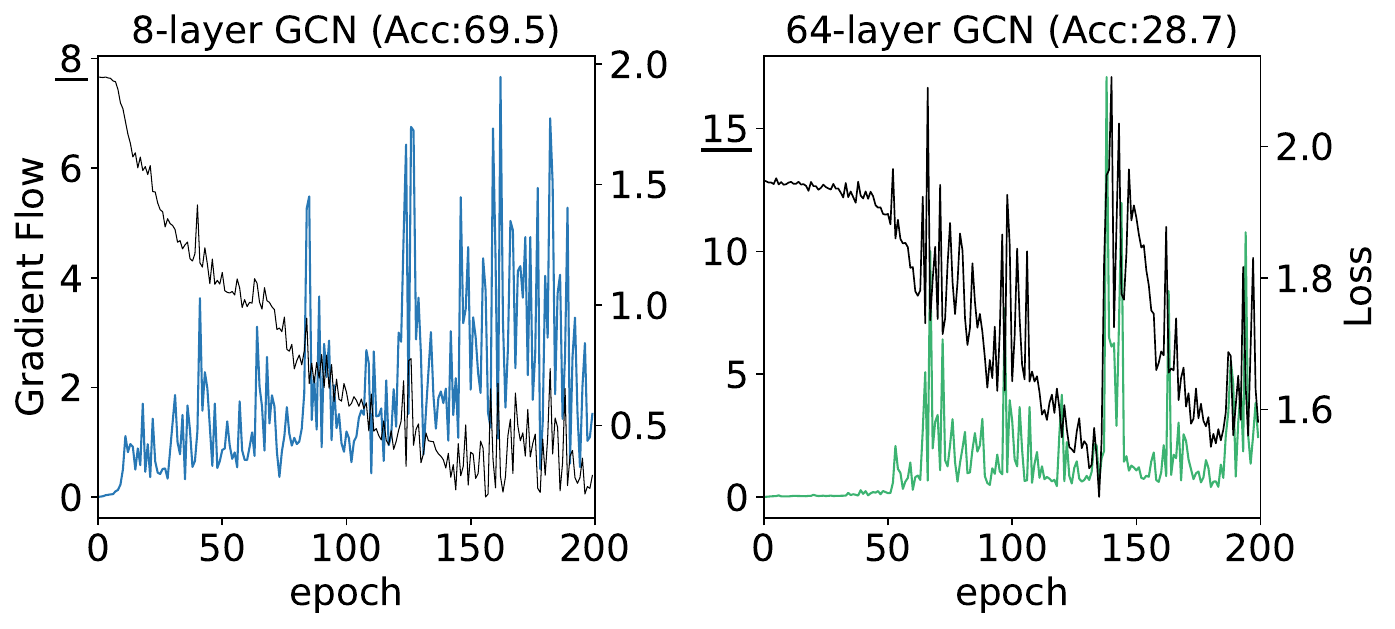}\hspace{3mm}
    \includegraphics[width=0.67\columnwidth]{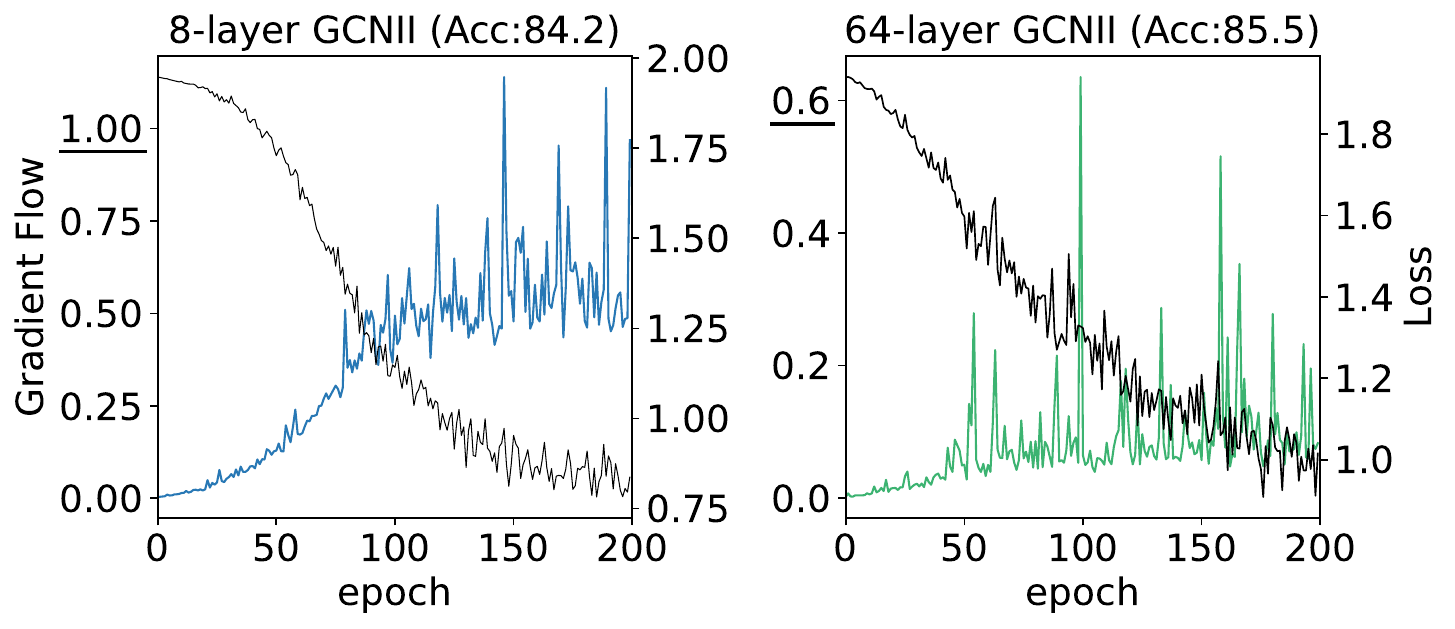}\\
    \vskip -0.10in
    \subfloat[ResGCN]{\includegraphics[width=0.67\columnwidth]{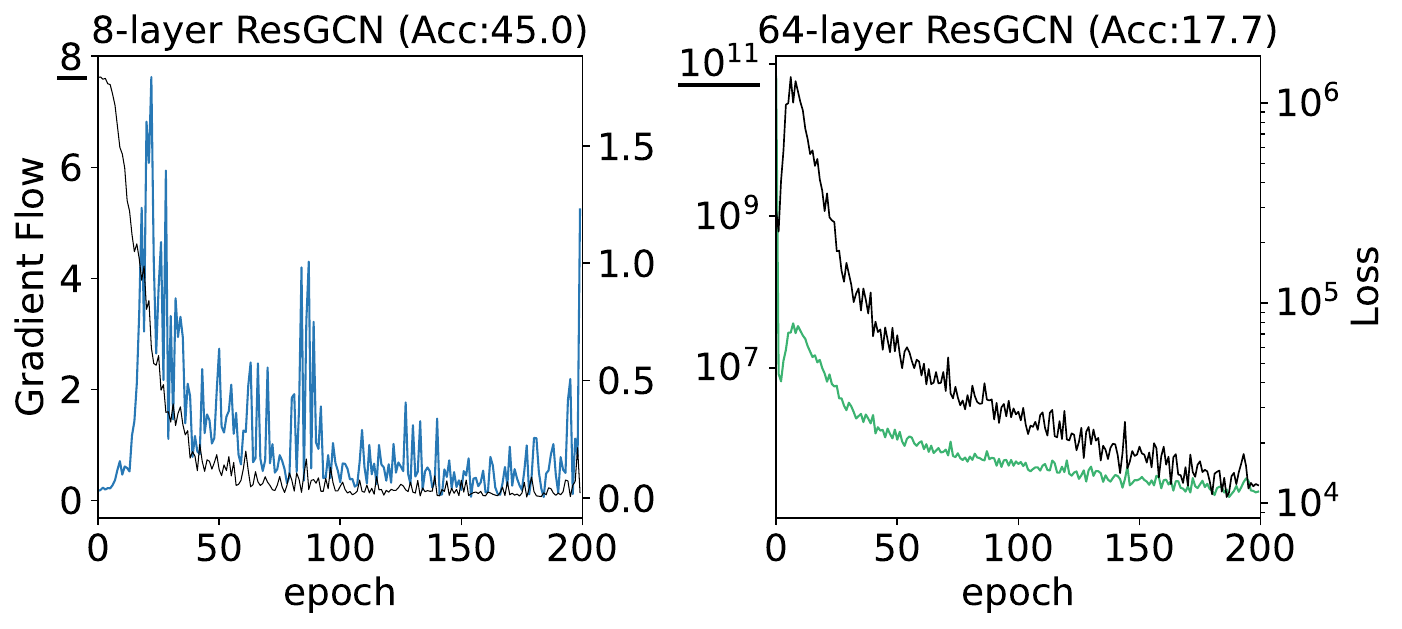}}\hspace{3.4mm}
    \subfloat[GCN]{\includegraphics[width=0.67\columnwidth]{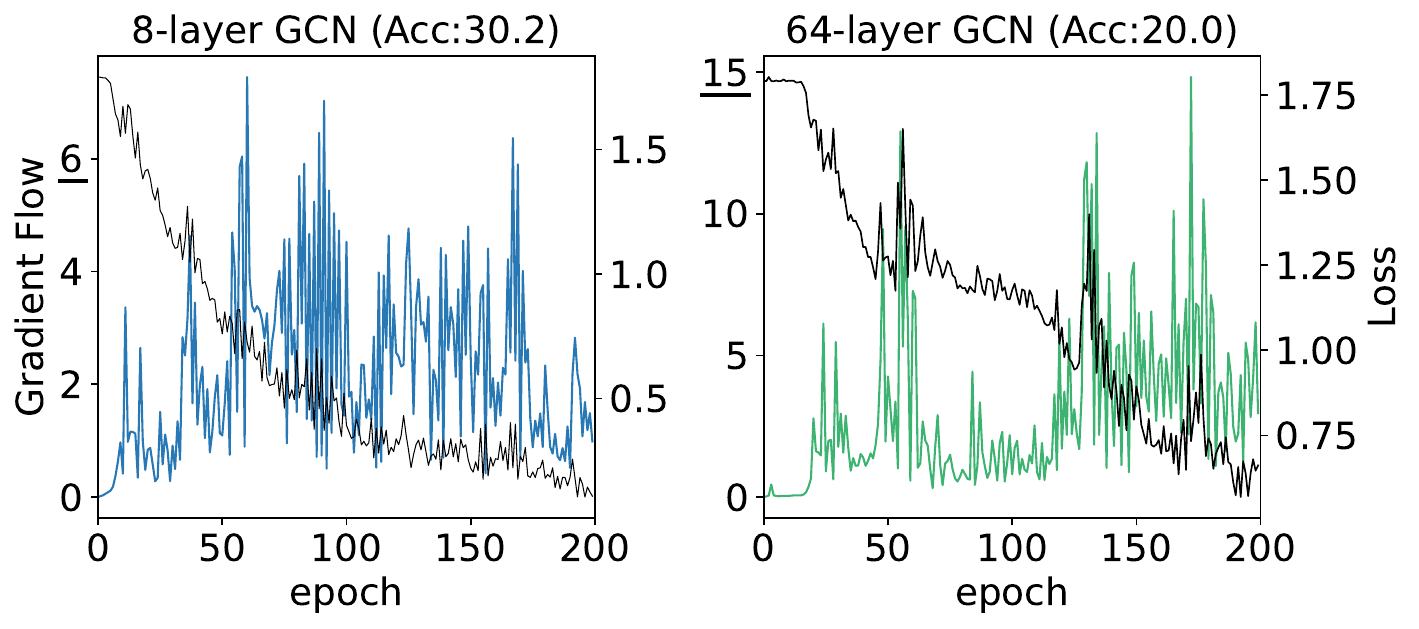}}\hspace{2.6mm}
    \subfloat[GCNII]{\includegraphics[width=0.67\columnwidth]{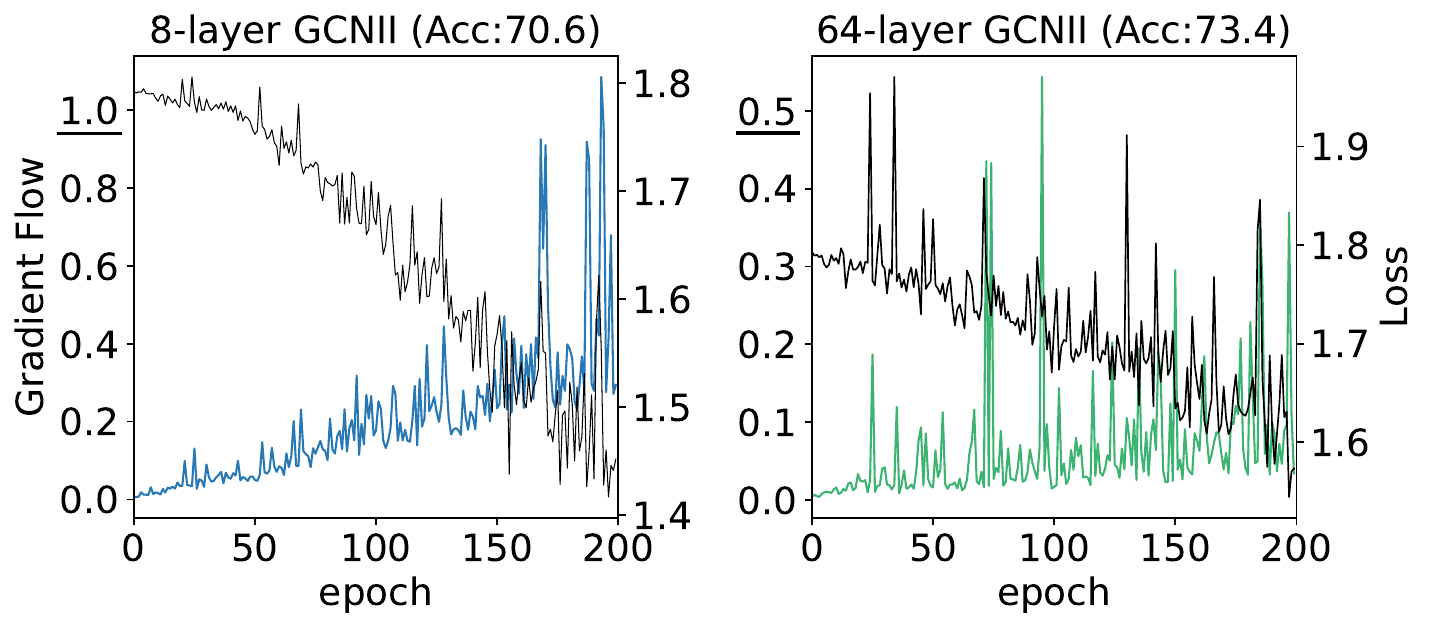}}\\
    \vskip -0.1in
    \caption{Gradient flow variation and loss in 8-layer and 64-layer ResGCN (a), GCN (b), and GCNII (c) trained on Cora (up) and Citeseer (down). \textbf{Blue} lines and \textbf{Green} lines denote the gradient flow of the 8-layer GNNs and that of the 64-layer GNNs, respectively. Black lines denote the loss variation during training. The corresponding test accuracy ($\%$) is given in the subtitle.}
    \label{fig:GFcora} 
\vskip -0.12in
\end{figure*}

\subsection{Deep Insight into Gradient Flow}\label{sec:deep insight}

The variation of the gradient flow is highly correlated with the trainability of the model, and the dramatic fluctuation of gradient flow is an unstable state in the learning dynamics of neural networks~\cite{glorot2010understanding,he2015delving,jaiswal2022old}.
In this section, we further investigate the state of the gradient flow as directly reflected by the gradient upper bounds of those GNN models. 
To be specific, we supplement the gradient upper bounds of ResGCN~\cite{li2019deepgcns}, a primitive Residual Connection for deep GNN, and $\omega$GCN~\cite{eliasof2023improving}, a state-of-the-art deep model with remarkable performance.
The results are shown in the last 2 rows of~\cref{tab:GNN_gradient bound}. 
For details of derivation, please refer to~\cref{extral_proofs}.

We focus on exploring the smoothing term and the training term in the gradient upper bounds of GCN, ResGCN, GCNII, and $\omega$GCN. 
For ResGCN, it is straightforward to find that the gradient upper bound of ResGCN is greater than that of GCN. 
In contrast, for GCNII, the parameter $\alpha<1$ diminishes the smoothing term.
Meanwhile, $\beta$, which decreases to zero with increased layer depth, also limits the training term. 
For $\omega$GCN, when the learnable parameter $\omega<1$, the gradient upper bound of $\omega$GCN will be lower than that of GCN as well, which is consistent with the practical learning results of $\omega$ in deep $\omega$GCN (shown in Figure 3 of~\citep{eliasof2023improving}). 
Hence, upon analyzing the structures of these bounds, we observe that the gradient upper bounds of GCN, ResGCN, GCNII, and $\omega$GCN satisfy the following order: ResGCN $>$ GCN $>$ GCNII / $\omega$GCN.

\textbf{A properly constrained smaller upper bound of the gradient leads to good trainability of deep GNNs.}
The gradient flow is defined as:  $\mathcal{GF}=\sum_{\ell=1}^{N}\left\|\frac{\partial \mathbf{L}(\mathbf{W})}{\partial \mathbf{W}^\ell}\right\|_2$. 
Hence, the gradient upper bound of each layer could directly affect the gradient flow upper bound of the corresponding deep GNN, reflecting the stability of the gradient flow. 
Given that the gradient upper bounds of GCN, ResGCN, GCNII, and $\omega$GCN follow the order mentioned above, the gradient flow upper bounds of them will also exhibit the identical order: $\mathcal{GF}($ResGCN$)>\mathcal{GF}($GCN$)>\mathcal{GF}($GCNII$)$ / $\mathcal{GF}(\omega$GCN$)$. 
Combined with the practical model performance shown in~\cref{fig:ACC}, a large gradient flow upper bound could lead to notable fluctuations of the gradient flow, limiting the trainability of ResGCN and GCN, thereby resulting in poor performance, especially when the number of layers goes deep.
On the contrary, the good performances of deep GCNII and $\omega$GCN illustrate that \emph{a more stable gradient flow (lower gradient flow upper bound) brought by a properly constrained smaller upper bound of the gradient }is one of the key reasons for the good trainability of deep networks. 
Meanwhile, exceedingly small gradient flows are not expected, which can lead to another training problem similar to the gradient vanishing.
The detailed experiments of gradient flow are conducted in~\cref{sec:exp}.

\section{Discussion on Constructing Deep Graph Models}
Based on our empirical experiments and theoretical gradient analysis, we clearly find that various existing methods that supposedly tackle Over-smoothing actually improve the trainability of MLPs, which is the main reason for their performance gains.
Consequently, we can concentrate on solving the trainability problem by regulating gradient flow without the distraction of the overstated Over-smoothing problem. 
Generally, there are two viable directions for further enhancing deep GNNs:
\begin{enumerate}
    \item Design a more theoretically explainable residual connection or introduce learnable parameters to control the graph propagation process.
    This approach aims to constrain \emph{the smoothing term}, indirectly improving both gradient flow and model trainability.
    
    \item Introduce parameters to control the norm of the weight matrix $\mathbf{W}$ or propose a new initialization strategy for $\mathbf{W}$.
    This method directly regulates \emph{the training term}, thereby leading to a stable gradient flow and good trainability.
\end{enumerate}

Beyond that, we doubt the feasibility of constructing deep GNNs under the current architecture. 
Although deeper models typically offer greater expressivity, it is crucial to weigh the challenges associated with deepening GNNs against the potential benefits, especially when considering most of the current graph scale. 
Because most of the existing graph scales are less than that of computer vision~\cite{aloysius2017review,khan2020survey}, language models~\cite{devlin2018bert,irie2019language} and other fields, which enjoy a lot from the deep models. 
Therefore, simply stacking deeper networks atop the existing architecture may not be the most effective approach for enhancing deep graph models. 
We propose that adjusting the paradigm of the graph model could avoid stacking numerous unnecessary linear layers for limited data.
Possible approaches include 1) reconstructing graph models with deep convolutional layers and shallow MLPs, 2) designing Transformer~\cite{vaswani2017attention} architecture tailored for graph-structured data, and 3) exploring novel paradigms with fewer training challenges in deep network scenarios.

\section{Experiment}\label{sec:exp}
In this section, we conduct experiments to verify the strong correlation between the gradient upper bound and the trainability for several deep GNNs, which is proved by our analysis above.

\textbf{Experimental Setup.}~~~We compare and analyze the variation in gradient flow of both shallow and deep GNNs in the training process of node classification tasks with three citation network datasets (Cora, Citeseer, and Pubmed) from~\citep{sen2008collective, yang2016revisiting}.
We use the standard training/validation/testing split of~\cite{yang2016revisiting} for citation network datasets in the semi-supervised scenario, with 20 nodes per class for training, 500 validation nodes, and 1,000 testing nodes. 
We focus on analyzing the gradient flow of ResGCN, GCN, and GCNII at various depths during training to validate the correctness of our theory. 
The formula for calculating the gradient flow is presented in~\cref{sec:deep insight}.

\textbf{Results.}~~~As witnessed in~\cref{fig:GFcora} to~\ref{fig:GFpub}, the gradient flows and the corresponding test accuracies of these models vary widely. 
On the one hand, \textbf{the peak values} (or the underlined values) of the gradient flow in ResGCN, GCN, and GCNII are consistent with our theoretical analysis in~\cref{sec:deep insight}, experimentally satisfying $\mathcal{GF}($ResGCN$)>\mathcal{GF}($GCN$)>\mathcal{GF}($GCNII$)$. 
On the other hand, for GNNs with properly small gradient flow upper bounds, such as GCNII, its gradient flow also maintains a stable update process even under the 64-layer network structure, which guarantees the model has good trainability and remarkable model performance.
On the contrary, for GNNs like GCN and ResGCN, their gradient flows fluctuate a lot, especially for ResGCN. 
This shows a problem similar to the gradient explosion, with gradient flow changing exponentially and resulting in poor model performance.
The above experimental results successfully verify our theoretical analysis and clarify that a properly constrained gradient upper bound can effectively improve the trainability of the model.
Meanwhile, the convergence tendency of loss in GCNII is more stable than that in GCN and ResGCN, which verifies the high correlation between the gradient upper bound and the model trainability as well.
See~\cref{app:exp gradient} for more results of gradient flow variation on other datasets.

\section{Conclusion}
In this work, we systematically analyze the dominant problem in deep GNNs and identify the issues that these deep GNNs towards addressing Over-smoothing actually work on via empirical experiments and theoretical gradient analysis.
We prove that Over-smoothing is not the dominant problem of deep GNNs and that the trainability issue of deep MLPs actually has a greater influence, which fills a theoretical gap in previous works.
Meanwhile, various existing methods that supposedly tackle Over-smoothing actually address the trainability issue, which is the main reason for their performance gains.
We further theoretically explore the gradient flow of deep GNNs, which illustrates that a properly constrained smaller upper bound of gradient flow is one of the key reasons for the good trainability of deep GNNs.
We conduct extensive experiments on various datasets to validate our theory.
Our analysis sheds light on constructing deep graph models in the future.

\begin{acks}
This research was supported in part by National Natural Science Foundation of China (No. U2241212, No. 61932001), by National Science and Technology Major Project (2022ZD0114802), by Beijing Natural Science Foundation (No. 4222028), by Beijing Outstanding Young Scientist Program No.BJJWZYJH012019100020098, and by Huawei-Renmin University joint program on Information Retrieval. 
We also wish to acknowledge the support provided by the fund for building world-class universities (disciplines) of Renmin University of China, by Engineering Research Center of Next-Generation Intelligent Search and Recommendation, Ministry of Education, by Intelligent Social Governance Interdisciplinary Platform, Major Innovation \& Planning Interdisciplinary Platform for the “Double-First Class” Initiative, by Public Policy and Decision-making Research Lab of Renmin University of China, and by Public Computing Cloud, Renmin University of China.
\end{acks}

\appendix
\begin{figure*}[t]
\vskip -0.2in
    \centering
    \subfloat[ResGCN]{\includegraphics[width=0.67\columnwidth]{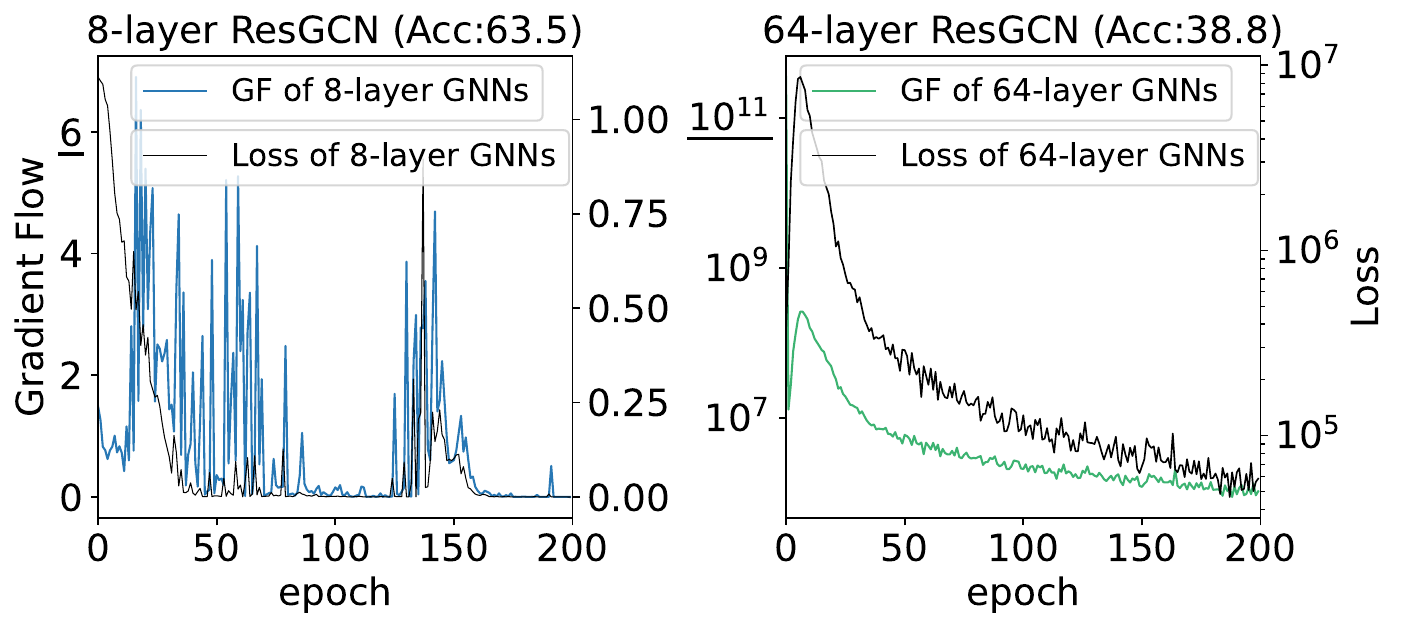}}\hspace{3mm}
    \subfloat[GCN]{\includegraphics[width=0.67\columnwidth]{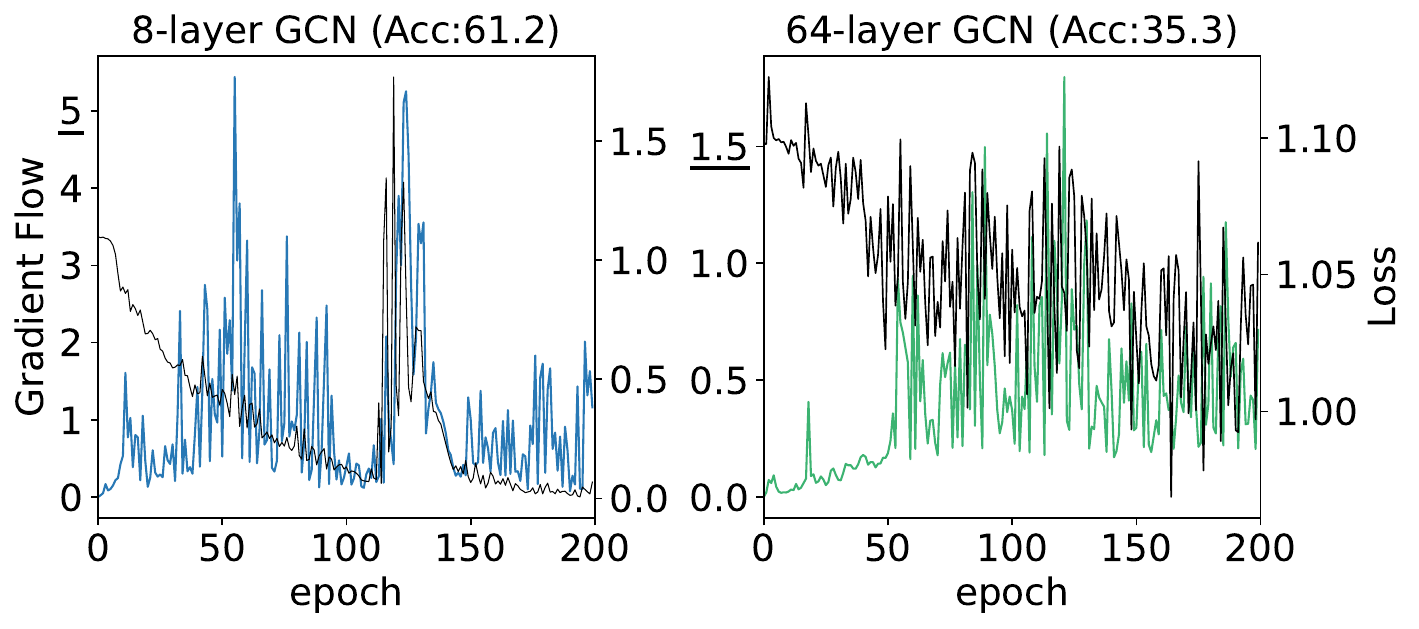}}\hspace{3mm}
    \subfloat[GCNII]{\includegraphics[width=0.67\columnwidth]{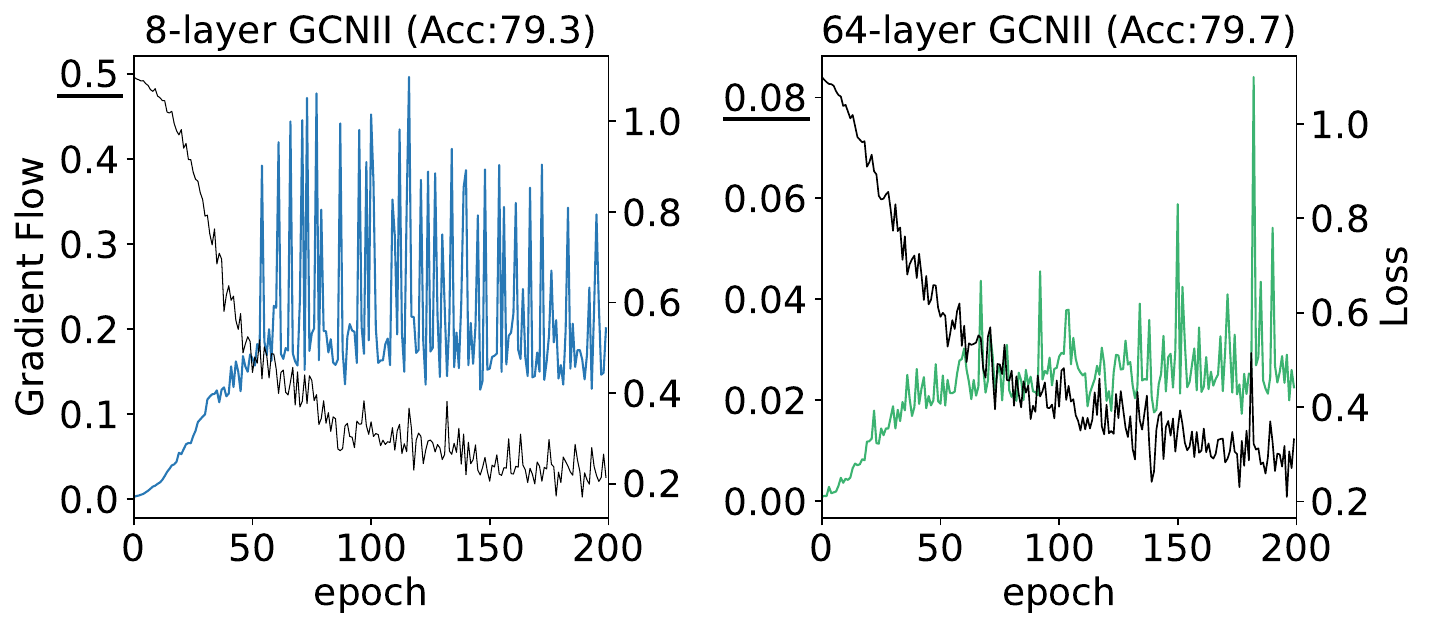}}\\
    \vskip -0.1in
    \caption{Gradient flow variation and loss in 8-layer and 64-layer ResGCN, GCN, and GCNII trained on Pubmed.}
    \label{fig:GFpub}
    \vskip -0.1in
\end{figure*}

\section{Proofs.}\label{app:proof}
\subsection{Proof of~\cref{pro:GCN graident bound}}\label{app:proof1}
$Proof.$ As witnessed in Equation~(\ref{equ:gradient calculation}), the calculation of $\frac{\partial\mathbf{L}(\mathbf{W})}{\partial{\mathbf{W}}^{(\ell)}_k}$ consists of three parts: $\frac{\partial\mathbf{L}(\mathbf{W})}{\partial{\mathbf{X}}^{(N)}}$, $\prod_{n=\ell+1}^{N}\frac{\partial\mathbf{X}^{(n)}}{\partial{\mathbf{X}}^{(n-1)}}$, and $\frac{\partial{\mathbf{X}}^{(\ell)}}{\partial{\mathbf{W}^{(\ell)}_k}}$. 
We derive and calculate these three parts one by one.

Firstly, the partial derivative of the first part $\frac{\partial\mathbf{L}(\mathbf{W})}{\partial{\mathbf{X}}^{(N)}}$ is calculated as:
\begin{equation}\label{equ:first part}
    \frac{\partial\mathbf{L}(\mathbf{W})}{\partial{\mathbf{X}}^{(N)}}=\frac{1}{v}\left[\mathbf{X}_{1}^{(N)}-{\mathbf{Y}}_{1},\mathbf{X}_{2}^{(N)}-{\mathbf{Y}}_{2},\cdots,\mathbf{X}_{v}^{(N)}-{\mathbf{Y}}_{v}\right].
\end{equation}

Then, the partial derivative of the second part $\prod_{n=\ell+1}^{N}\frac{\partial\mathbf{X}^{(n)}}{\partial{\mathbf{X}}^{(n-1)}}$ is calculated as: $\prod_{n=\ell+1}^{N}\frac{\partial\mathbf{X}^{(n)}}{\partial{\mathbf{X}}^{(n-1)}}=\prod_{n=\ell+1}^{N}\Phi^{(n)}$, 
where $\Phi^{(n)}\in\mathbb{R}^{v\times v}$. 
The detailed structure is as follows:
\begin{equation}
\begin{aligned}
    \Phi^{(n)}&=\left[\frac{\partial\mathbf{X}^{(n)}}{\partial\mathbf{X}^{(n-1)}_1},\frac{\partial\mathbf{X}^{(n)}}{\partial\mathbf{X}^{(n-1)}_2},...,\frac{\partial\mathbf{X}^{(n)}}{\partial\mathbf{X}^{(n-1)}_v}\right]\\
    &=\begin{bmatrix}\frac{\partial\mathbf{X}_1^{(n)}}{\partial\mathbf{X}_1^{(n-1)}},&\cdots&,\frac{\partial\mathbf{X}_1^{(n)}}{\partial\mathbf{X}_v^{(n-1)}}\\\vdots&&\vdots\\\frac{\partial\mathbf{X}_v^{(n)}}{\partial\mathbf{X}_1^{(n-1)}},&\cdots&,\frac{\partial\mathbf{X}_v^{(n)}}{\partial{X}_v^{(n-1)}}\end{bmatrix}.
\end{aligned}
\end{equation}

Based on~\cref{equ:node-wise GCN}, for all $1\leq i\leq v$, it is easy to have:
\begin{equation}\label{equ:second part}
\begin{aligned}
    \Phi^{(n)}_{i,i}&=\sigma^{\prime}\cdot\frac{\mathbf{W}_i^{(n)}}{d_i},\\
    \Phi^{(n)}_{i,j}&=\sigma^{\prime}\cdot\frac{\mathbf{W}_j^{(n)}}{\sqrt{d_id_j}},~~j\in\mathcal{N}_i,\\
    \Phi^{(n)}_{i,j}&=0,~~j\notin\mathcal{N}_{i}\mathrm{~and~}j\neq i.
\end{aligned}
\end{equation}

Finally, the partial derivative of the third part $\frac{\partial{\mathbf{X}}^{(\ell)}}{\partial{\mathbf{W}^{(\ell)}_k}}=\Psi^{(\ell)}$,
where $\Psi^{(\ell)}\in\mathbb{R}^{v}$.
The detailed element information is as follows:
\begin{equation}
\begin{aligned}
    \Psi^{(\ell)}=\left[\frac{\partial\mathbf{X}^{(\ell)}_1}{\partial\mathbf{W}^{(\ell)}_k},\frac{\partial\mathbf{X}^{(\ell)}_2}{\partial\mathbf{W}^{(\ell)}_k},...,\frac{\partial\mathbf{X}^{(\ell)}_v}{\partial\mathbf{W}^{(\ell)}_k}\right].
\end{aligned}
\end{equation}

Based on~\cref{equ:node-wise GCN}, for all $1\leq i\leq v$, it is easy to have:
\begin{equation}\label{equ:third part}
\begin{aligned}    
    \Psi^{(\ell)}_i&=\sigma^{\prime}\cdot\frac{\mathbf{X}_i^{(\ell-1)}}{d_i},~~i=k,\\
    \Psi^{(\ell)}_i&=\sigma^{\prime}\cdot\frac{\mathbf{X}_k^{(\ell-1)}}{\sqrt{d_id_k}},~~k\in\mathcal{N}_i,\\
    \Psi^{(\ell)}_i&=0,~~k\notin\mathcal{N}_i\mathrm{~and~}i\neq k.
\end{aligned}
\end{equation}

After the precise calculation of the above partial derivatives, we try to find an upper bound of the gradient at each update.
Based on Equation~(\ref{equ:first part}),~(\ref{equ:second part}) and~(\ref{equ:third part}), it is straightforward to obtain the upper bound of each part as follows:
\begin{equation}
\begin{aligned}
    \left\|\frac{\partial\mathbf{L}(\mathbf{W})}{\partial{\mathbf{X}}^{(N)}}\right\|_\infty&\leq\frac{1}{v}\left(\max_{1\leq i\leq v}\left|\mathbf{X}_i^{(N)}\right|+\max_{1\leq i\leq v}\left|{\mathbf{Y}}_i\right|\right),\\
    \left\|\frac{\partial\mathbf{X}^{(n)}}{\partial{\mathbf{X}}^{(n-1)}}\right\|_\infty&\leq \max\left(\sigma^\prime\right)\cdot\max_{1\leq i\leq v,j\in\mathcal{N}_i}\frac{1}{\sqrt{d_id_j}}\cdot\left\|\mathbf{W}^{(n)}\right\|_\infty,\\
    \left\|\frac{\partial{\mathbf{X}}^{(\ell)}}{\partial{\mathbf{W}^{(\ell)}_k}}\right\|_\infty&\leq\max\left(\sigma^\prime\right)\cdot\max_{1\leq i\leq v,j\in\mathcal{N}_i}\frac{1}{\sqrt{d_id_j}}\cdot\left\|\mathbf{X}^{(\ell-1)}\right\|_\infty.
\end{aligned}
\end{equation}

Hence, the node-wise gradient of deep GCN with regards to any learnable weight parameter ${\mathbf{W}}^{(\ell)}_k$, for $1\leq k\leq v$, $1\leq\ell\leq N$ is bounded as:
\begin{equation}
\begin{aligned}
    \left|\frac{\partial\mathbf{L}(\mathbf{W})}{\partial{\mathbf{W}}^{(\ell)}_k}\right|
    \leq&\frac{1}{v}\left(\max_{1\leq i\leq v}\left|\mathbf{X}_i^{(N)}\right|+\max_{1\leq i\leq v}\left|{\mathbf{Y}}_i\right|\right)\cdot\\
    &\left(\max_{1\leq i\leq v,j\in\mathcal{N}_i}\frac{1}{\sqrt{d_id_j}}\right)^{N-\ell+1}\cdot\prod_{n=\ell+1}^{N}\left\|\mathbf{W}^{(n)}\right\|_\infty\cdot\left\|\mathbf{X}^{(\ell-1)}\right\|_\infty.
\end{aligned}
\end{equation}

The proof of~\cref{pro:GCN graident bound} ends.

\subsection{Proofs of the Bounds in~\cref{tab:GNN_gradient bound}}\label{extral_proofs}
The main idea of the proofs is similar to the proof of~\cref{pro:GCN graident bound} in~\cref{app:proof1}.

\section{Results of Decoupled Experiments}\label{app:decouple}
\begin{figure}[h]
    \centering
    \includegraphics[width=\columnwidth]{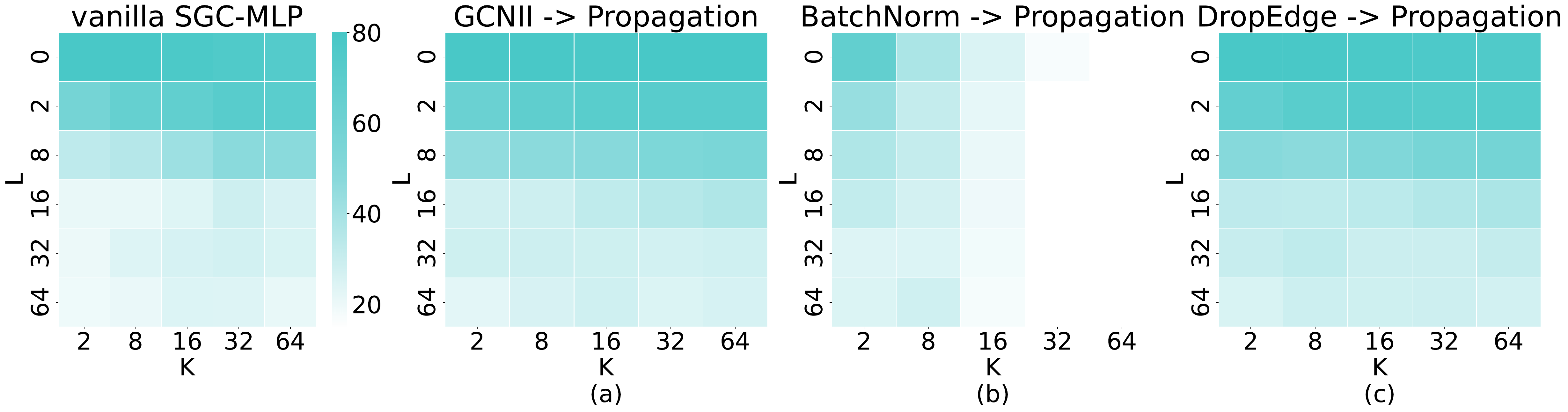}
    \vskip -0.05in
    \caption{The results of decoupled experiments of node classification tasks on Cora. Batch Normalization is applied to the graph propagation process for experimental completeness.}
    \label{fig:BN}  
    \vskip -0.05in
\end{figure}
Consistent with the analysis in~\cref{sec:revisit exp}, adding Batch Normalization to the graph propagation process is not functional. According to the performance of Batch Normalization in~\cref{fig:BN}(b), the introduction of Batch Normalization to the propagation process even further exacerbates the performance degradation of the model.

\begin{figure}[h]
    \centering
    \includegraphics[width=\columnwidth]{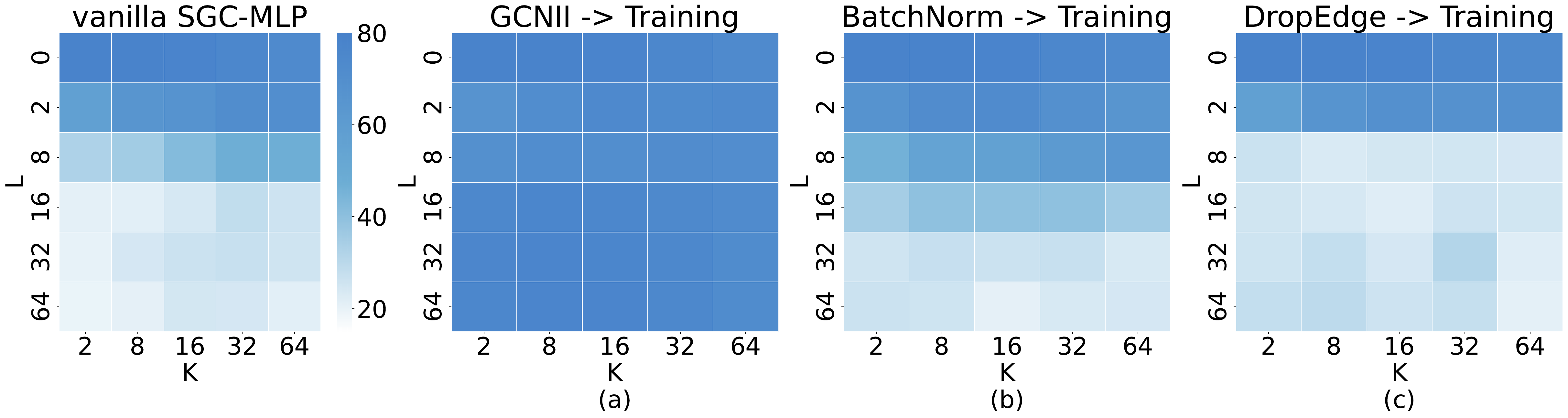}
    \vskip -0.05in
    \caption{The results of decoupled experiments of node classification tasks on Cora. For experimental completeness, we use DropOut~\cite{srivastava2014dropout} with a ratio of 60$\%$ to approximate the effect of DropEdge during training since they can be essentially viewed as Drop operations in different dimensions~\cite{fang2023dropmessage}}
    \label{fig:DE}  
    \vskip -0.05in
\end{figure}
Although we try to approximate DropEdge with other Drop operations in the training process, such as DropOut~\cite{srivastava2014dropout}, the result in~\cref{fig:DE}(c) shows that adding Drop operations to the training process does not improve model performance. 
Meanwhile, DropOut is actually a common model optimization strategy. 
Therefore, consistent with the analysis in~\cref{sec:revisit exp}, DropEdge cannot directly play a role in optimizing the training process.

\section{Additional experimental results}\label{app:exp gradient}
The results of the gradient flow experiments on Pubmed are shown in~\cref{fig:GFpub}, which are consistent with our theoretical analysis in~\cref{sec:deep insight} as well. 

\newpage
\bibliographystyle{ACM-Reference-Format}

\bibliography{reference}



\end{document}